\documentclass[12pt]{article}

\usepackage{macros}

\usepackage[margin = .95in]{geometry}

\pdfoutput=1

\usepackage{amsmath}

\usepackage{graphicx}

\usepackage{amssymb,amsmath}

\usepackage{relsize}

\usepackage{upgreek}

\usepackage{subfigure}

\usepackage{url}

\usepackage{graphicx}

\usepackage{algorithmic}
\usepackage{algorithm2e}

\usepackage{xcolor}

\usepackage{amsopn}

\DeclareMathOperator{\argmin}{\mathrm{argmin}}

\newcommand{\diam}[1]{\mathrm{Diam}(#1)}

\def\mmu{\mathlarger{\mathlarger{\upmu}}}

\def\M{\mathcal{M}}

\def\X{\mathbf{X}}

\def\part{\frac{\partial}{\partial \X_{i,j}}}

\def\E{\mathbf{E}}

\def\v{\mathbf{v}}

\def\R{\mathbb{R}}

\def\e{\mathbf{e}}

\def\B{\mathcal{B}}

\def\W{\mathbf{W}}

\usepackage{multirow}

\newcommand{\mcl}[2]{\multicolumn{#1}{c}{#2}}

\title{Degrees of Freedom and Model Selection for $k$-means Clustering}

\begin{document}

\author{\name David P.\ Hofmeyr \hfill
{\small \textmd{ Department of Statistics and Actuarial Science}}\\
\textcolor{white}{.}\hfill {\small \textmd{Stellenbosch University}}\\
\textcolor{white}{.}\hfill {\small \textmd{7600, South Africa}}
}

\maketitle

\begin{abstract}%
  This paper investigates the model degrees of freedom in $k$-means clustering. An extension of Stein's lemma provides an expression for the effective degrees of freedom in the $k$-means model. Approximating the degrees of freedom in practice requires simplifications of this expression, however empirical studies evince the appropriateness of our proposed approach. The practical relevance of this new degrees of freedom formulation for $k$-means is demonstrated through model selection using the Bayesian Information Criterion. The reliability of this method is validated through experiments on simulated data as well as on a large collection of publicly available benchmark data sets from diverse application areas. Comparisons with popular existing techniques indicate that this approach is extremely competitive for selecting high quality clustering solutions. Code to implement the proposed approach is available in the form of an {\tt R} package from \url{https://github.com/DavidHofmeyr/edfkmeans}.
\end{abstract}


%% -- Introduction -------------------------------------------------------------

%% - In principle "as usual".
%% - But should typically have some discussion of both _software_ and _methods_.
%% - Use {}, \pkg{}, and \code{} markup throughout the manuscript.
%% - If such markup is in (sub)section titles, a plain text version has to be
%%   added as well.
%% - All software mentioned should be properly \citep-d.
%% - All abbreviations should be introduced.
%% - Unless the expansions of abbreviations are proper names (like "Journal
%%   of Statistical Software" above) they should be in sentence case (like
%%   "generalized linear models" below).

\section{Introduction}

Degrees of freedom arise explicitly in model selection, as a way of accounting for the bias in the model log-likelihood for estimating generalisation performance~\citep[Akaike Information Criterion, AIC]{akaike1998information} and, indirectly, Bayes factors~\citep[Bayesian Information Criterion, BIC]{schwarz1978estimating}. In particular, degrees of freedom account for the complexity, or flexibility of a model by measuring its effective number of parameters.
In the context of clustering, model flexibility is varied primarily by different choices of $k$, the number of clusters. In $k$-means, clusters are associated with compact collections of points arising around a set of {\em cluster centroids}. The optimal centroids %specifically the optimal centroids
%, $\pmu_1, ..., \pmu_k$, 
are those which minimise the sum of squared distances between each point and its assigned centroid. Using the squared distance connects the $k$-means objective with the log-likelihood of a simple Gaussian Mixture Model (GMM).
% of the Gaussian Mixture Model (GMM).
%The optimal $k$-means solution may be seen as an approximation of the maximum likelihood solution for a GMM in which the covariance matrices and mixing proportions are equal, and the covariance matrices are proportional to the identity. 
Pairing elements of the GMM log-likelihood with AIC and BIC type penalties, based on the number of explicitly estimated parameters, has motivated multiple model selection methods for $k$-means~\citep{manning,ramsey2008uncovering,xmeans}. However, it has been observed that these approaches can lead to substantial over-estimation of the number of clusters~\citep{hamerly2004learning}.

We argue that these simple penalties are inappropriate, and do not account for the entire complexity of the model, and investigate more rigorously the degrees of freedom in the $k$-means model. The proposed formulation depends not only on the explicit dimension of the model, but also accounts for the uncertainty in the cluster assignments. This is intuitively appealing, as it allows the degrees of freedom to incorporate the difficulty of the clustering problem, which cannot be captured solely by the model dimension. 
%For example, the greater the degree of cluster overlap, the more challenging the clustering problem and the more uncertain the solution. This results in a greater complexity needed to fit to the underlying distribution.
% In the remainder the model degrees of freedom are investigated and a new formulation is presented.
This formulation draws on the work of~\cite{TibshiraniSUREExtend}, and is the first application, of which we are aware, of this approach to the problem of clustering. We validate the proposed formulation by applying it within the BIC to perform model selection for $k$-means. The approach is found to be extremely competitive with the state-of-the-art on a very large collection of benchmark data sets.

The remaining paper is organised as follows. In Section~\ref{sec:dfmodel} we discuss the $k$-means model explicitly, and consider its degrees of freedom. We also provide details for how we approximate the degrees of freedom practically. Section~\ref{sec:BIC} describes our approach for model selection based on the Bayesian Information Criterion and using these approximated degrees of freedom. Section~\ref{sec:experiments} documents the results from a thorough simulation study as well as comparisons between the proposed approach and popular existing methods on simulated data, as well as on a very large collection of publicly available benchmark data sets. Finally, we give some concluding remarks in Section~\ref{sec:discussion}.
 
 %of this section is devoted to methodology for obtaining an estimate of the model degrees of freedom in the $k$-means solution.% Multiple interpretations of both the model and likelihood components are reasonable, as are their relationships with the model degrees of freedom. After discussion of these different interpretations, a new approach for estimating the model degrees of freedom is provided.
% Thereafter the proposed method is compared against popular existing techniques on both simulated data and data from real applications.

\section{Degrees of Freedom in the $k$-means Model} \label{sec:dfmodel}

%Degrees of freedom arise explicitly in model selection, as a way of accounting for the bias in the model log-likelihood for estimating generalisation error (AIC) and, indirectly, Bayes factors (BIC).
%In model selection, degrees of freedom arise explicitly as a means to account for the bias in the model log-likelihood for estimating generalisation error (AIC) and, indirectly, Bayes factors (BIC).
From a probabilistic perspective, the standard modelling assumptions for $k$-means are that the data arose from a $k$ component Gaussian mixture in $d$ dimensions with equal isotropic covariance matrix, $\sigma^2 I$, and either equal mixing proportions~\citep{manning, celeux1992classification} or sufficiently small $\sigma$~\citep{jiang2012small}. In this case, and with a slight abuse of notation, one may in general write the likelihood for the data, given model $\M$, which we assume to include all parameters of the underlying distribution which are being estimated, as
\begin{align*}
\ell(\X|\M) %&= \sum_{i=1}^n\log\left(\sum_{j=1}^k P(\X_{i\_}\in C_j)f\left(\X_{i\_}| \X_{i\_}\in C_j\right)\right)\\
&= \sum_{i=1}^n\log\left(\sum_{j=1}^k\pi_{ij} \frac{1}{(2\pi\sigma^2)^{d/2}} \exp\left(-\frac{||\X_{i\_} - \pmb{\mu}_j||^2}{2\sigma^2}\right)\right).
%&= \sum_{i=1}^n\log\left(\frac{1}{k(2\pi)^{d/2}\sigma^d}\sum_{j=1}^k \exp\left(-\frac{\sum_{l=1}^d(\X_{il} - \mmu_{jl})^2}{2\sigma^2}\right)\right),
\end{align*}
%
%Here $C_1, ..., C_k$ are the mixture components (clusters),
Here $\pmb{\mu}_1, ..., \pmb{\mu}_k \in \R^d$ are the component means, $\pi_{ij}$ is the probability that the $i$-th datum arises from the $j$-th component, and the subscript ``$i\_$'' is used to denote the $i$-th row of a matrix. The terms $\pi_{ij}$ are usually assumed equal for fixed $j$, and have been used to represent mixing proportions~\citep{manning}. %However, as we discuss below, this has frequently been applied in an inconsistent way. 
% Here we mean for $\M$ to include all elements of the underlying distribution which are being estimated.
%
%The Bayesian Information Criterion is popular in model selection, and may be defined as
%
%\begin{align*}
%\mathrm{BIC}(\X; \M) = -2\ell(\X|\M) + \log(n)\mathrm{df}(\M),
%\end{align*}
%
%where $\X\in \R^{n\times d}$ is the observed data matrix, $\ell(\X|\M)$ is the log-likelihood of the data, given a specific model $\M$, and $\mathrm{df}(\M)$ is the model degrees of freedom. From a probabilistic perspective, the standard modelling assumptions for $k$-means are that the data arose from a $k$ component Gaussian mixture with equal isotropic covariance matrix, $\sigma^2 I$, and either equal mixing proportions~\citep{manning, celeux1992classification} or sufficiently small $\sigma$~\citep{jiang2012small}.
%
%it is common (GET REFERENCES) to assume that
%$k$-means modelling assumptions, 
%the data arose from a $k$ component Gaussian mixture with equal mixing proportions and equal isotropic covariance matrices. 
Popular formulations of the $k$-means likelihood~\citep{manning,ramsey2008uncovering,xmeans} use the so-called classification likelihood~\citep{fraley2002model}, which treats the cluster assignments as true class labels. For example, a simple BIC formulation has been expressed, up to an additive constant, as~\citep{ramsey2008uncovering}
\begin{align}\label{eq:bic_xmeans}
\frac{1}{\sigma^2}\sum_{i=1}^n\min_{j\in \{1,...,k\}}||\X_{i\_} - \hat{\pmb{\mu}}_j||^2 + \log(n)kd.
\end{align}
Here only the means are assumed part of the estimation, and hence the model dimension is $kd$, for $k$ clusters. There is a fundamental mismatch in formulations such as this, however, including those in~\cite{manning,ramsey2008uncovering,xmeans}, between the log-likelihood component and the bias correction term.
Specifically, by using the classification likelihood the assumption is that the model is also estimating the assignments of data to clusters.
%
%Most importantly, however, this formulation ignores the contribution to the likelihood for the $i$-th datum from all but the component whose estimated mean is nearest. There is thus an implicit assumption that the model also estimates the assignment of data to clusters.
%
%We may interpret this as having the model estimate the probabilities $\pi_{ij}$ under the constraint that for each $i$ all but one $\pi_{ij}$ are zero. 
However, without incorporating this added estimation into the model degrees of freedom, the bias of the log-likelihood for estimating generalisation error, and Bayes factors, is severely under-estimated.
%this added estimation must be accommodated in the model degrees of freedom. 
%Notice that correcting this is not as simple as adding $n$ degrees of freedom, one for each assignment, since the value of the means and the assignment of data to clusters are not independent.

%Some authors have considered more appropriate formulations of the BIC~\citep[e.g.]{xmeans}, which include the estimation of the scale and mixing proportions. As far as we are aware, however, no authors have considered the assignments of data to clusters in the estimation of model degrees of freedom.

In this work a modified formulation is considered which incorporates the cluster assignment into the modelling procedure. We find it convenient to assume that the data matrix $\X$ has been generated as,
\begin{align}\label{eq:generative}
\X = \mmu + \E,
\end{align}
where the mean matrix $\mmu \in \R^{n\times d}$ is assumed to have $k$ unique rows  and the elements of $\E \in \R^{n\times d}$ are independent realisations from a $N(0, \sigma^2)$ distribution. Notice that in this case the log-likelihood may be written as,
\begin{align*}
\ell(\X | \mmu, \sigma) &= \sum_{i=1}^n \sum_{j=1}^d \log\left(\frac{1}{\sqrt{2\pi}\sigma}\exp\left(-\frac{(\X_{i,j}-\mmu_{i,j})^2}{2\sigma^2}\right)\right)\\
&= -\frac{1}{2\sigma^2}\sum_{i=1}^n \sum_{j=1}^n (\X_{i,j} - \mmu_{i,j})^2 - nd \log(\sigma) + K,
\end{align*}
for constant $K$ independent of $\sigma$ and $\mmu$. Note that in this formulation the assignment of data (rows of $\X$) to mixture components is captured implicitly by the $k$ distinct rows of $\mmu$. Also notice that if $\sigma$ is assumed fixed then this is essentially equivalent (up to an additive constant) to the likelihood term in the BIC formulation in~(\ref{eq:bic_xmeans}) above. % Now, depending on whether or not the variance term, $\sigma^2$, forms part of the estimation, we find that

For this formulation it is possible to consider estimating pointwise the elements of $\mmu$, under the constraint of having $k$ unique rows, using a modelling procedure $\M:\R^{n\times d}\to \R^{n\times d}$, defined as
% which assigns each row of $\X$ to the cluster centroid to which it is nearest, and the centroids minimise the sum of the squared distances between the data and their assigned centroid. That is,
% :\X \mapsto \hat{\mmu}$, where $\hat{\mmu}$ is the estimated mean matrix, given by
%
\begin{align}
\M(\X)_{i,j} &= \hat{\mmu}_{c(i), j} \label{eq:model1}\\
\hat{\mmu} &= \argmin_{\mathbf{M}\in \R^{k\times d}}\sum_{i=1}^n \min_{l\in\{1, ..., k\}}||\X_{i\_}-\mathbf{M}_{l\_}||^2 \label{eq:model2}\\
c(i) &= \argmin_{l\in\{1, ..., k\}}||\X_{i\_} - \hat{\mmu}_{l\_}||^2. \label{eq:model3}
\end{align}
The matrix $\hat{\mmu} \in \R^{k\times d}$ estimates the unique rows of $\mmu$, and provides an approximation of the maximum likelihood solution under Eq~(\ref{eq:generative}). The indices $c(i), i = 1, ..., n$ indicate the assignments of the data (rows of $\X$) to the different clusters' means (rows of $\hat \mmu$). %In the next subsection the degrees of freedom consumed in the above modelling procedure will be investigated.
With this formulation we are able address the estimation of the ``effective degrees of freedom''~\citep{efron1986biased}, given by
\begin{align}\label{eq:generaliseddf}
\mathrm{df}(\M) = \frac{1}{\sigma^2}\sum_{i=1}^n\sum_{j=1}^d \mbox{Cov}(\M(\X)_{i,j}, \X_{i,j}).
\end{align}
The covariance offers an appealing interpretation in terms of model complexity/flexibility. A more complex model will respond more to variations in the data, in that additional flexibility will allow the model to attempt to ``explain'' this variation. The covariance between its fitted values and the data will therefore be higher. On the other hand, an inflexible model will, by definition, vary less due to changes in the observations. Furthermore in numerous simple Gaussian error models there is an exact equality between this covariance and the model dimension.
The remainder of this section is concerned with obtaining an appropriate approximation of the effective degrees of freedom for the $k$-means model. The following two lemmas are useful for obtaining such an estimate.

%\begin{figure}
%\centering
%\subfigure[Degrees of Freedom]{\includegraphics[width = .45\textwidth]{eg1df.pdf}\label{fig:eg1df}}
%\subfigure[Bayesian Information Criterion]{\includegraphics[width = .45\textwidth]{eg1BIC.pdf}\label{fig:eg1bic}}
%\includegraphics[height = .5\textwidth, width = .8\textwidth]{edfVdfBIC.pdf}
%\caption{Degrees of freedom and approximate BIC for 10 component Gaussian mixture in 10 dimensions. Standard $\mathrm{df}$ estimate (- - - -) and estimate based on proposed appraoch ($-\circ-\circ-$)\label{fig:eg1}}
%\end{figure}

\begin{lemma}\label{lem:lem1}

Let $\X = \mmu + \E \in \R^{n\times d}$, with $\mmu$ fixed and $\E_{i,j}\sim N(0, \sigma^2)$ with $\E_{i,j}, \E_{k,l}$ independent for all $(i,j) \not = (l,k)$. Let $f:\R^{n\times d}\to \R^{n\times d}$ %be such that $\bigcup_{i=1}^n\{f(\W)_{i\_}\} \subset \mbox{\bf conv}\left(\bigcup_{j=1}^n \{\W_{j\_}\}\right)$ for all $\W \in \R^{n\times d}$, where {\bf conv}$(\cdot)$ denotes the convex hull. Assume also that $f$
satisfy the following condition. For all $\W \in \R^{n\times d}$ and each $i,j$, %and $\{\W_{k,l}\}_{(k,l)\not = (i,j)}$,
 there exists a finite set $\mathcal{D}^\W_{i,j} = \bigcup_{l=1}^q\{\delta_l\}$ s.t. $f$, viewed as a univariate function by keeping all other elements of $\W$, $\{\W_{k,l}\}_{(k,l)\not = (i,j)}$, fixed, is Lipschitz on each of $(-\infty, \delta_1), (\delta_1, \delta_2), ..., (\delta_{q-1}, \delta_q)$, and $(\delta_q, \infty)$. Then for each $i,j$, the quantity $\frac{1}{\sigma^2}Cov(f(\X)_{i,j}, \X_{i,j})$ is equal to
\begin{align}\label{eq:lem1}
E\left[\frac{\partial}{\partial \X_{i,j}}f(\X)_{i,j}\right]
+ \frac{1}{\sigma}E\Bigg[\sum_{\delta: \X_{i,j}+\delta \in \mathcal{D}_{i,j}^\X}\phi\left(\frac{\X_{i,j}+\delta - \mmu_{i,j}}{\sigma}\right)
\lim_{\gamma \downarrow\uparrow \delta}f(\X+\gamma \e_{i,j})_{i,j}\Bigg],
\end{align}
provided the second term on the right hand side exists. Here $\phi(x) = (2\pi)^{-1/2}\exp(-x^2/2)$ is the Gaussian density function; $\e_{i,j}\in\R^{n\times d}$ has zero entries except in the $i,j$-th position, where it takes the value one; and
$$
\lim_{\gamma \downarrow\uparrow \delta}f(\X+\gamma \e_{i,j}) = \lim_{\gamma\downarrow \delta}f(\X+\gamma \e_{i,j}) - \lim_{\gamma\uparrow \delta}f(\X+\gamma \e_{i,j}).
$$
is the size of the discontinuity at $\delta$.

\end{lemma}

\noindent

This result is very similar to~\cite[Lemma 5]{TibshiraniSUREExtend}, where the regression context is considered. Our proof is given in the appendix. The first term in~(\ref{eq:lem1}) comes from Stein's influential result~\citep[Lemma 2]{SteinSURE} for determining the risk in the estimation of the mean of a Gaussian random variable using a smooth model. 
Due to the discontinuities in the $k$-means model, which occur at points where the cluster assignments of some of the data change, the additional covariance at the discontinuity points needs to be accounted for. Consider an $\X$ which is close to a point of discontinuity with respect to the $i,j$-th entry. Conditional on the fact that $\X$ is close to such a point, $f(\X)_{i,j}$ takes values approximately equal to the left and right limits, depending on whether $\X_{i,j}$ is below or above the discontinuity respectively. On a small enough scale each happens with roughly equal probability. After taking into account the probability of being close to the discontinuity point, and taking the limit as $\X$ gets arbitrarily close to the discontinuity point, one can arrive at an intuitive justification for the additional term in~(\ref{eq:lem1}). In the remainder this additional covariance term will be referred to as the excess degrees of freedom.

In the above result the function $f$ may be seen to represent an arbitrary modelling procedure, which takes as argument a data matrix and outputs a matrix of fitted values which represent an estimate of the means of the elements in the data under a Gaussian error model. The next lemma places Lemma~\ref{lem:lem1} in the context of the $k$-means model, where it is verified that the modelling procedure $\M$, described in Eqs.~(\ref{eq:model1})--(\ref{eq:model3}), satisfies the conditions described above. Notice that in this context, the discontinuities in the model (the $\delta$ values in the statement of Lemma~\ref{lem:lem1}) correspond with the points at which some of the clustering assignments would change.

\begin{lemma}

Let $\M:\R^{n\times d}\to \R^{n\times d}$ be defined as
\begin{align*}
\M(\W)_{i,j} = \hat{\mmu}_{c(i), j},
\end{align*}
where
\begin{align*}
\hat{\mmu} &= \argmin_{\mathbf{M}\in \R^{k\times d}}\sum_{i=1}^n \min_{j\in\{1, ..., k\}}||\W_{i\_}-\mathbf{M}_{j\_}||^2\\
c(i) &= \argmin_{j\in\{1, ..., k\}}||\W_{i\_} - \hat{\mmu}_{j\_}||^2.
\end{align*}
Then $\M$ satisfies the conditions on the function $f$ in the statement of Lemma~\ref{lem:lem1}, and moreover if $\X = \mmu + \E \in \R^{n\times d}$, with $\mmu$ fixed and $\E_{i,j}\sim N(0, \sigma^2)$ with $\E_{i,j}, \E_{k,l}$ independent for all $(i,j) \not = (l,k)$, then
\begin{align*}
E\Bigg[\sum_{\delta: \X_{i,j}+\delta \in \mathcal{D}^\X_{i,j}}\phi\left(\frac{\X_{i,j}+\delta - \mmu_{i,j}}{\sigma}\right)\lim_{\gamma \downarrow\uparrow \delta}\M(\X+\gamma \e_{i,j})_{i,j}\Bigg]
\end{align*}
exists and is finite.

\end{lemma}

\noindent
One of the most important consequences of~\citep[Lemma 2]{SteinSURE}, which leads to the first term in~(\ref{eq:lem1}), is that this term is devoid of any of the parameters of the underlying distribution. An unbiased estimate of this term can be obtained by taking the partial derivatives of the model using the observed data. In the case of $k$-means one arrives at,
\begin{align*}
\frac{\partial \M(\X)_{i,j}}{\partial \X_{i,j}} = \frac{\partial \hat \mmu_{c(i),j}}{\partial \X_{i,j}} = \frac{1}{n_{c(i)}},
\end{align*}
where $n_{c(i)}$ is the number of data assigned to centroid $c(i)$. Therefore,
\thinmuskip = 1mu
\begin{align*}
\sum_{i=1}^n\sum_{j=1}^d \frac{\partial \M(\X)_{i,j}}{\partial \X_{i,j}} &= \sum_{j=1}^d \sum_{l=1}^k \sum_{i: c(i) = l}\frac{\partial \M(\X)_{i,j}}{\partial \X_{i,j}}\\
&= \sum_{j=1}^d \sum_{l=1}^k \sum_{i: c(i) = l}\frac{1}{n_{c(i)}}=\sum_{j=1}^d \sum_{l=1}^k n_{c(i)}\frac{1}{n_{c(i)}} = kd.
\end{align*}
The excess degrees of freedom therefore equals the difference between the effective degrees of freedom and the explicit model dimension, i.e., the number of elements in $\hat \mmu$. It may therefore be interpreted as the additional complexity in assigning data to clusters. This is intuitively pleasing in light of the fact that this additional covariance directly accounts for the potential assignment of the data to different clusters, in that these are what result in discontinuities in the model.

%(REDO THIS PIECE) Figure~\ref{fig:eg1} shows plots of the degrees of freedom and corresponding BIC values, using the estimated effective degrees of freedom, and also the commonly adopted $\mathrm{df} = kd$. The data arise from a Gaussian mixture containing ten components. The inter-class distances are roughly equal and the overlap is relatively large to illustrate the effect of ambiguity in the solution when $k$ is incorrectly specified. The left panel of Figure~\ref{fig:eg1} shows plots of the degrees of freedom for varying $k$. Although perhaps counter-intuitive in a more general setting, the effective degrees of freedom decreases for increasing $k$, for the few values leading up to the correct number of 10, whereafter it increases as $k$ exceeds the correct number. This effect translates to the BIC, where in the right panel of Figure~\ref{fig:eg1} there is a clear and unique minimum at $k=10$ and where the standard approach would have selected the largest value of $k$ considered.\\
% Important to note too is that the effective degrees of freedom estimates are greater than the standard estimate for all values of $k$.\\
% This exceedance of the standard estimate has been consistently observed across all experiments conducted for this research.\\
%\\

\subsection{Approximating Excess Degrees of Freedom}

The excess degrees of freedom reintroduces the unknown parameters to the degrees of freedom expression. %, and hence invariably one must use estimates. 
Furthermore, as noted by~\cite{TibshiraniSUREExtend}, it is generally extremely difficult to determine the discontinuity points, making the computation of the excess degrees of freedom very challenging. This perhaps even more so in the case of clustering. %These points are discussed below in further detail.
 Consider the excess degrees of freedom arising from the $i,j$-th entry,
\begin{align*}
&\frac{1}{\sigma}E\Bigg[\sum_{\delta: \X_{i,j}+\delta \in \mathcal{D}^\X_{i,j}}\phi\left(\frac{\X_{i,j}+\delta - \mmu_{i,j}}{\sigma}\right)\lim_{\gamma \downarrow\uparrow \delta}\M(\X+\gamma \e_{i,j})_{i,j}\Bigg].
\end{align*}
%
%Since the observed data matrix, $\X$, is assumed to have come from the distribution of $\Y$, the excess degrees of freedom are estimated using $\X$ and the corresponding clustering solution. 
Assume for now that the model parameters, $\mmu$ and $\sigma^2$, are fixed. We will discuss our approach for accommodating these unknown parameters in the next subsection. % will be replaced with the natural candidate, using the rows of $\hat \mmu$, and $\sigma^2$ will for now be assumed known.
Now, recall that the discontinuities $\mathcal{D}^\X_{i,j}$ are those $\delta$ at which the assignment of some of the data changes. That is, those $\delta$ for which $\exists m$ s.t.
\begin{align*}
\lim_{\gamma \downarrow\uparrow \delta}\argmin_{l = 1, ..., k} ||(\X + \gamma \e_{i,j})_{m\_} - \hat\mmu(\X+\gamma \e_{i,j})_{l\_}|| \not = 0,
\end{align*}
%
%where we make explicit the dependence of the estimated mean, $\hat \mmu$, on the data.
%where $\e_{i,j}$ is as before the matrix with zeros except the $i,j$-th entry where it takes the value one. 
%Here the dependence of $\hat\mmu$ on the data matrix is stressed explicitly since the location and magnitude of the discontinuities are not determined in relation to the observed matrix, $\X$, but rather the modified matrix, $\X + \delta \e_{i,j}$. 
The fact that discontinuities are determined in terms of the would-be solution, $\hat\mmu(\X+\gamma \e_{i,j})$, rather than the observed solution, $\hat\mmu(\X)$, is one of the reasons % It is this fact 
which make determining the discontinuity points extremely challenging. Here we have made explicit the dependence of the estimated means, $\hat \mmu$, on the data. Indeed, one can construct examples where slight changes in only a single matrix entry can result in reassignments of arbitrarily large subsets of data, resulting in substantial and unpredictable changes in $\hat\mmu$.
%Admittedly limited preliminary experiments have led to the conclusion that in simplified cases the contributions to the excess degrees of freedom arising from discontinuities in the fitted value of $\X_{i,j}$, which come about as a result of a reassingment of another row of $\X$, are negligible, in general contributing less than $0.5\%$ of the total.
We are thus led to making some simplifications. First, we only consider discontinuities w.r.t. the $i,j$-th entry arising from reassignments of $\X_{i\_}$, the corresponding datum. This is a necessary simplification which maintains the intuitive interpretation of the excess degrees of freedom as the covariance arising from reassignments of data. %, while inducing only moderate bias in simple cases. %To approximate the bias in the general setting seems at this stage an insurmountable task.
Now, consider the value of $\delta$ at which the assignment of $\X_{i\_}$ changes from $c(i)$ to some $l \not = c(i)$. Ignoring all other clusters, we find that $\delta$ satisfies
%It thus becomes necessary to make simplifying assumptions. Specifically it is assumed that only values of $\delta$ for which 
%\begin{align*}
%\underline \delta (i):=\inf\{\gamma < 0 | c(i)_\gamma = c(i)\} \leq \delta \leq \sup\{\gamma > 0 | c(i)_\gamma = c(i)\}=: \overline \delta (i),\\ c(i)_\gamma := \argmin_{l = 1, ..., k} ||\X + \gamma \e_{j} - \hat\mmu(\X + \gamma \e_{i,j})_{l\_}||,
%\end{align*}
%where $\e_j$ is the $j$-th canonical basis vector, result in a substantial contribution to the excess degrees of freedom. That is, the values of $\delta$ between where $\X_{i\_}$ would first be reassigned by either increasing or decreasing $\delta$ from zero. This assumption is justified by the fact that for $\delta$ large enough in magnitude to be a point of discontinuity {\em after} $\X_{i\_}$ has already moved far enough to be reassigned to another cluster, the term $\phi\left(\frac{\X_{i,j}+\delta - \hat\mmu_{i,j}}{\sigma}\right)\lim_{\gamma \downarrow\uparrow \delta}\M(\X+\gamma \e_{i,j})_{i,j}$ is likely to be extremely small. Now, to obtain estimates of these bounds, $\underline \delta (i)$ and $\overline \delta (i)$ it is necessary to determine the points at which $c(i)_\delta$ changes from $c(i)$ to $l$ for each $l \not = c(i)$, while ignoring the changes of other points and the assignment of $\X_{i\_}$ to other clusters. The largest negative, and the smallest positive such values provide initial estimates. Now, the point at which $c(i)_\delta$ changes from $c(i)$ to $l$ must satisfy,
%
\begin{align}\label{eq:deltaval}
\left\|\X_{i\_} + \delta \e_{j} - \hat \mmu_{c(i)\_} - \frac{\delta}{n_{c(i)}}\e_j\right\|^2 &= \left\|\X_{i\_} + \delta \e_{j} - \hat \mmu_{l\_}\right\|^2,
\end{align}
where $\e_j$ is the $j$-th canonical basis vector for $\R^d$ and $n_{c(i)}$ is the size of the $c(i)$-th cluster. This is a quadratic equation which can easily be solved. A further simplification is adopted here. Rather than considering the paths of $\X_{i\_}$ through multiple reassignments resulting from varying $\delta$ (which quickly become extremely difficult to calculate), the magnitude and location of a discontinuity at a value $\delta$ is determined as though no reassignments had occurred for values between zero and $\delta$. 
%It is fairly straightforward to reason that this will tend to lead to a positive bias in the total excess degrees of freedom. However, %
Since the corresponding values of $\delta$ are generally large, the contributions from the quantities $\phi\left(\frac{\X_{i,j}+\delta - \mmu_{i,j}}{\sigma}\right)\lim_{\gamma \downarrow\uparrow \delta}\M(\X+\gamma \e_{i,j})_{i,j}$ are generally small, and hence we expect the bias induced by this simplification to be relatively small.
The excess degrees of freedom for the $i,j$-th entry is thus approximated using
\begin{align}\label{eq:dfij}
%&\sum_{\delta: \X_{i,j}+\delta \in \mathcal{D}^\X_{i,j}}\frac{1}{\sigma}\phi\left(\frac{\X_{i,j}+\delta - \mmu_{i,j}}{\sigma}\right)\lim_{\gamma \downarrow\uparrow \delta}\M(\X+\gamma \e_{i,j})_{i,j}\\
%& \hspace{30pt} \approx 
\frac{1}{\sigma}\sum_{l \not = c(i)}\phi\bigg(\frac{\X_{i,j}+\delta_l-\mmu_{i,j}}{\sigma}\bigg)\lim_{\gamma \downarrow\uparrow \delta_l}\M(\X+\gamma \e_{i,j})_{i,j},
%\left(\hat\mmu_{c(i),j}-\frac{n_l}{n_l+1}\hat\mmu_{l,j}-\frac{\X_{i,j}}{n_l+1}+\delta_l\left(\frac{n_l+1 - n_{c(i)}}{n_{c(i)}(n_l+1)}\right)\right)\\
%& - \sum_{\substack{l \not = c(i)\\ \delta_l > 0}}\frac{1}{\sigma}\phi\bigg(\frac{\X_{i,j}+\delta_l-\hat\mmu_{c(i),j}}{\sigma}\bigg)\left(\hat\mmu_{c(i),j}-\frac{n_l}{n_l+1}\hat\mmu_{l,j}-\frac{\X_{i,j}}{n_l+1}+\delta_l\left(\frac{n_l+1 - n_{c(i)}}{n_{c(i)}(n_l+1)}\right)\right),
\end{align}
where $\delta_l$ is the solution to Eq.~(\ref{eq:deltaval}) with smaller magnitude (when a solution exists). To determine the magnitude of the discontinuities observe that when $\delta_l<0$, and we assume, as above, that no values $\delta_l < \delta < 0$ result in a reassignment of $\X_{i\_}$, we have
\begin{align}
\nonumber
&\lim_{\gamma \downarrow \delta_l}\M(\X + \gamma \e_{i,j})_{i,j} = \hat\mmu_{c(i),j} + \frac{\delta_l}{n_{c(i)}},\\
&
\nonumber
\lim_{\gamma \uparrow \delta_l}\M(\X + \gamma \e_{i,j})_{i,j} = \frac{1}{n_l+1}\left(n_l\hat\mmu_{l,j} + \X_{i,j} + \delta_l\right)\\
&\Rightarrow \lim_{\gamma \downarrow\uparrow \delta_l}\M(\X + \gamma \e_{i,j})_{i,j} = \hat\mmu_{c(i),j}-\frac{n_l}{n_l+1}\hat\mmu_{l,j}-\frac{\X_{i,j}}{n_l+1}+\delta_l\left(\frac{n_l+1 - n_{c(i)}}{n_{c(i)}(n_l+1)}\right)\label{eq:discont}.
\end{align}
If $\delta_l>0$ then we simply have the negative of the above.
% Next the selection of $\sigma$ will be discussed.
%
%Now, notice that the locations and magnitudes of the discontinuities in $\M(\X)_{i,j}$ do not depend on $\sigma$.
% For each $i,j$ let $\mathcal{D}_{i,j}$ be the locations and $\mathcal{J}_{i,j}$ the magnitudes of the discontinuities described above, and let $\mathcal{D} = \bigcup_{i,j} \mathcal{D}_{i,j}$ and $\mathcal{J} = \bigcup_{i,j}\mathcal{J}_{i,j}$.
%To obtain an estimate, $\hat \sigma$, to be used in computing the degrees of freedom, a straightforward solve-the-equation approach is employed. That is, if we define
%
%\begin{align*}
%\hat \sigma(\mathrm{df}) = \frac{\sum_{i=1}^n ||\X_{i\_} - \hat \mmu_{c(i)\_}||^2}{nd - \mathrm{df}},
%\end{align*}
%
%then $\hat\sigma$ is the solution which satisfies
%
%\begin{align*}
%\mathrm{df} &= kd + \frac{1}{\hat\sigma(\mathrm{df})}\sum_{i=1}^n\sum_{j=1}^d\sum_{\substack{l \not = c(i)}}\phi\bigg(\frac{\X_{i,j}+\delta^{i,j}_l-\hat\mmu_{c(i),j}}{\hat\sigma(\mathrm{df})}\bigg)\mathcal{J}_{i,j,l},\\
%\mathcal{J}_{i,j,l}&:= -\mathrm{sign}(\delta_l^{i,j})\left(\hat\mmu_{c(i),j}-\frac{n_l}{n_l+1}\hat\mmu_{l,j}-\frac{\X_{i,j}}{n_l+1}+\delta^{i,j}_l\left(\frac{n_l+1 - n_{c(i)}}{n_{c(i)}(n_l+1)}\right)\right),
%\end{align*}
%
%where $\delta_l^{i,j}$ is as $\delta_l$ above, but now with an explicit dependence on the indices $i,j$. In the author's experience only one or two iterations are required for convergence of the value of $\mathrm{df}$ in the above.

\subsubsection{Selecting Appropriate Values for {\Large $\upmu$} and $\sigma^2$ for Estimating Degrees of Freedom}

The estimate of excess degrees of freedom depends on the values of $\mmu$ and $\sigma^2$. It is tempting to use the apparently natural candidates, based on $\hat\mmu$ and an estimate of the within cluster variance from the model, whose degrees of freedom are being estimated, itself. However, this is inappropriate for the purpose of comparing models. First, notice that the value of $\hat\mmu$ will lead to an underestimation of the terms, $\phi\left(\frac{\X_{i,j}+\delta-\mmu_{i,j}}{\sigma}\right)$. This is because the values which result in a reassignment of the corresponding datum occur at the boundaries of the estimated clusters; and hence, on average, at the greatest distances from $\hat\mmu$. Furthermore, note that smaller values of $\sigma^2$ tend to result in a smaller value of the estimated degrees of freedom, everything else being equal. A model with an over-estimation of $k$ would lead to an underestimation of $\sigma^2$, and hence an artifically low estimated degrees of freedom. Such a model would thus be penalised insufficiently, relatively to those with a smaller number of clusters, and hence larger estimate of $\sigma^2$.
%
%
%It is therefore preferable that the same set of values of $\sigma^2$ are used for all values of $k$, the number of clusters. Otherwise the effect is that a hypothesis of a larger number of clusters tends to be self-fulfilling. This is because the within cluster variance estimate (an apparently sensible value for $\sigma^2$) from a large number of clusters will be relatively small, decreasing the excess degrees of freedom artificially, and hence penalising models with a large number of clusters insufficiently, relatively to those with a smaller number of clusters, and hence larger estimate of $\sigma^2$.

We have observed that to estimate the degrees of freedom for a model with $k$ clusters, a reasonable approximation can often be obtained by using the estimated parameters from any larger model (i.e., one with a greater number of clusters). In particular, if we now let $\M(\X;k)$ be the fitted values from Eqs.~(\ref{eq:model1})--(\ref{eq:model3}), making explicit the number of clusters in the model, then replacing $\mmu$ and $\sigma$ with $\M(\X;k')$ and $\sqrt{\frac{1}{nd}\sum_{i=1}^n\sum_{j=1}^d (\X_{i,j} - \M(\X;k')_{i,j})^2}$ respectively, where $k' > k$, provides a reasonable estimate of the degrees of freedom in model $\M(\X; k)$. It is interesting that the estimate of degrees of freedom is similar for a large range of values $k'$, provided they are greater than $k$. Let's consider again the terms in the excess degrees of freedom, i.e., terms of the form
\begin{align*}
\frac{1}{\sigma} \phi\left(\frac{\X_{i,j} + \delta - \mmu_{i,j}}{\sigma}\right)\lim_{\gamma \uparrow \downarrow \delta} \M(\X + \gamma \mathbf{e}_{i,j}; k)_{i,j}.
\end{align*}
Now, notice that the term inside $\phi$ may be seen as having two components, namely $\frac{\X_{i,j} - \mmu_{i,j}}{\sigma}$ and $\frac{\delta}{\sigma}$. The first of these will tend to be similar for different $k'$ when a complementary pair of $\mmu$ and $\sigma$ is used. Indeed, replacing these with the estimates described above, averaging their squared values over all $i, j$ produces a constant, independent of $k'$. Furthermore, notice that, in general, $\delta$ will have the same sign as $\X_{i,j} - \mmu_{i,j}$, since $\delta$ is the value which causes a change in the assignment of the datum $\X_{i\_}$ from its nearest cluster mean. The term $\phi\left(\frac{\X_{i,j} + \delta - \mmu_{i,j}}{\sigma}\right)$ will therefore tend to decrease, in general, when considering all pairs $i, j$, as $k'$ increases. Conveniently, this decrease is approximately counteracted by the fact that the terms in the excess degrees of freedom include the factor $1/\sigma$, which increases as $k'$ increases.

\begin{figure}[h]
\subfigure[Wine Data Set]{\includegraphics[width = \textwidth]{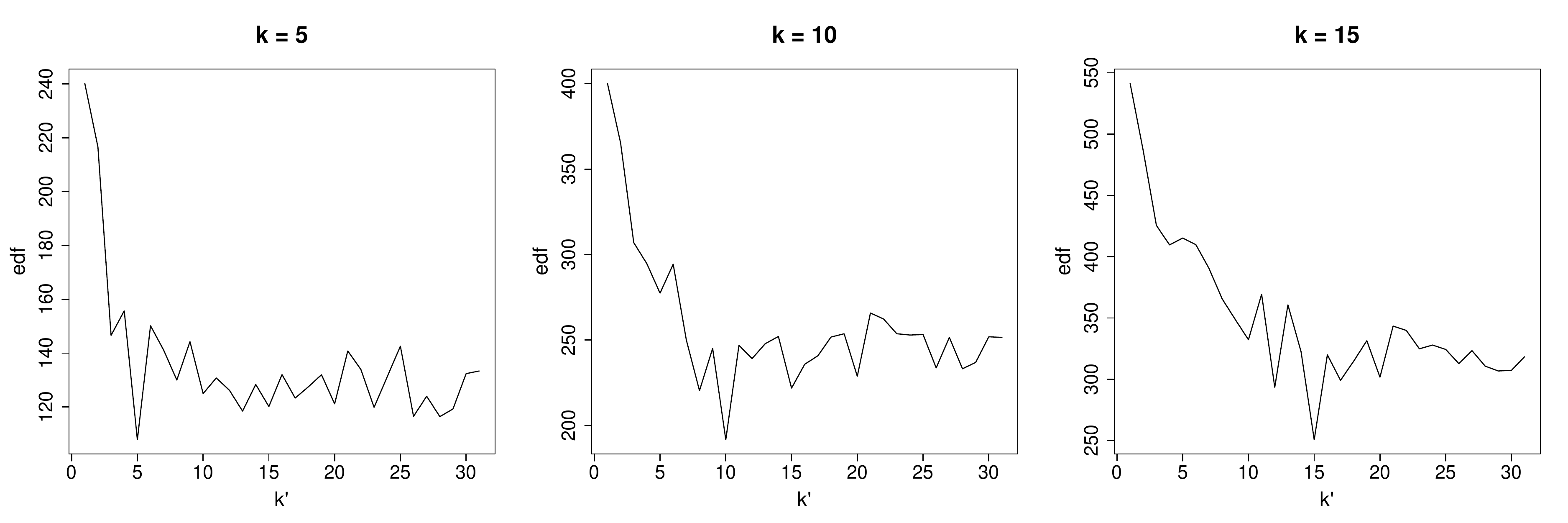}}
\subfigure[Optical Recognition of Handwritten Digits Data Set]{\includegraphics[width = \textwidth]{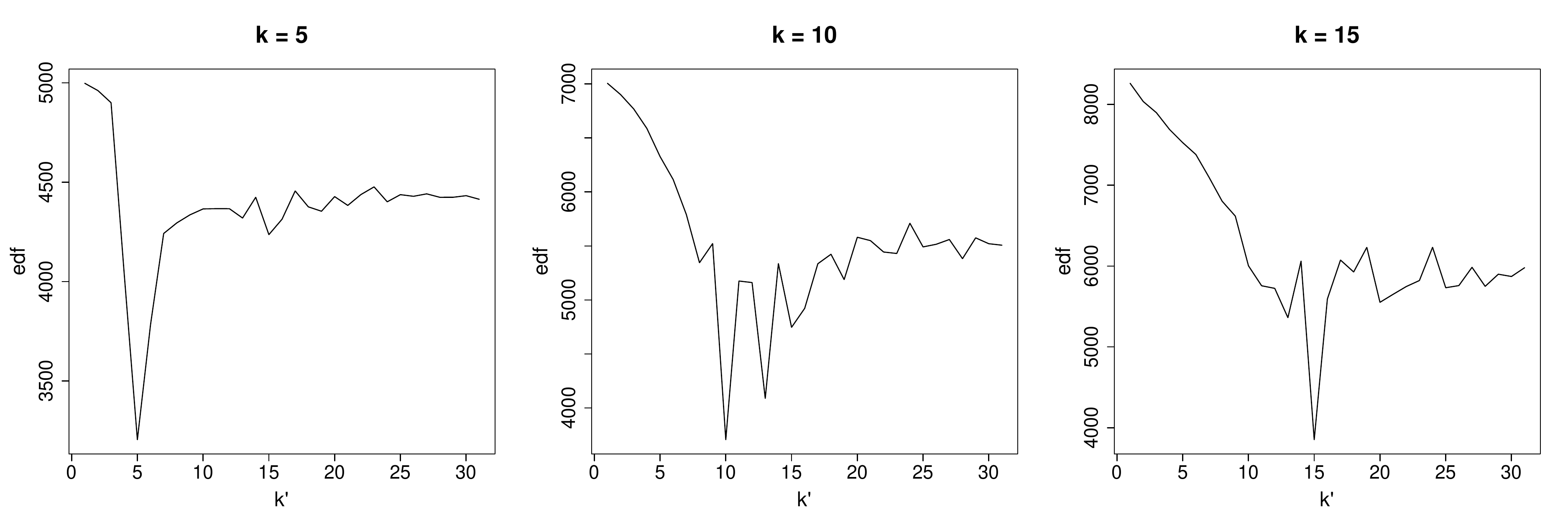}}
\caption{Estimated degrees of freedom for models with $k = 5, 10$ and 15 clusters, using the estimated parameters from models with a number of clusters, $k'$, ranging from 1 to 30. There is a clear dip in the estimates when $k = k'$, but the estimates are stable for $k' > k$. \label{fig:edfs}}
\end{figure}

Figure~\ref{fig:edfs} shows the estimated degrees of freedom from $k$-means models obtained from two of the data sets used in our applications\footnote{both data sets are available from the UCI machine learning repository~\citep{UCI}}. For each of the two data sets we have shown the estimated degrees of freedom for the models with 5, 10 and 15 clusters, and for varying $k'$. There is a very clear dip in the plots where $k = k'$, caused by underestimation of the degrees of freedom by replacing the unknown parameters with the estimates from the same model. As described above, however, the estimates then become stable for values $k' > k$. In practice we simply set $k' = k_{max} + 1$, where $k_{max}$ is the largest number of clusters under consideration, to estimate the degrees of freedom for all values of $k$.

\subsection{Accuracy of the Approximated Degrees of Freedom}

Here we briefly report on a short set of simulations designed to assess the accuracy of the degrees of freedom approximation we have introduced. To begin, we quickly recap our approach.
To approximate the degrees of freedom in the model $\M(\X; k)$, i.e., the $k$-means solution with $k$ clusters, we first compute, for each $i, j$, those values of $\delta$ at which datum $\X_{i\_}$ would be assigned to another cluster, if shifted in direction $\mathbf{e}_j$. That is, for each $l\not = c(i)$, we compute $\delta_l^{i,j}$ according to Eq.~(\ref{eq:deltaval}), where we have now introduced explicitly into the notation the indices $i,j$. We then compute the sizes of the model discontinuities at these values of $\delta$, i.e., the values $\lim_{\gamma \downarrow\uparrow \delta^{i,j}_l}\M(\X + \gamma \mathbf{e}_{i,j})_{i,j}$, using Eq.~(\ref{eq:discont}). Finally, we set
\begin{align}\label{eq:edf_final}
\widehat{\mathrm{df}(\M(\X; k))} = \frac{1}{\tilde\sigma}\sum_{i=1}^n\sum_{j=1}^d\sum_{l\not = c(i)} \phi\left(\frac{\X_{i,j}+\delta_l^{i,j}-\tilde{\mmu}_{i,j}}{\tilde\sigma}\right)\lim_{\gamma \downarrow\uparrow \delta^{i,j}_l}\M(\X + \gamma \mathbf{e}_{i,j})_{i,j},
\end{align}
where $\tilde{\mmu} = \M(\X; k')$ and $\tilde\sigma = \sqrt{\frac{1}{nd}\sum_{i=1}^n\sum_{j=1}^d (\X_{i,j} - \tilde{\mmu}_{i,j})^2}$ for some $k' > k$. 

\begin{figure}
\centering
\subfigure[5 clusters in 5 dimensions]{\includegraphics[width = 3.9cm, height = 2.8cm]{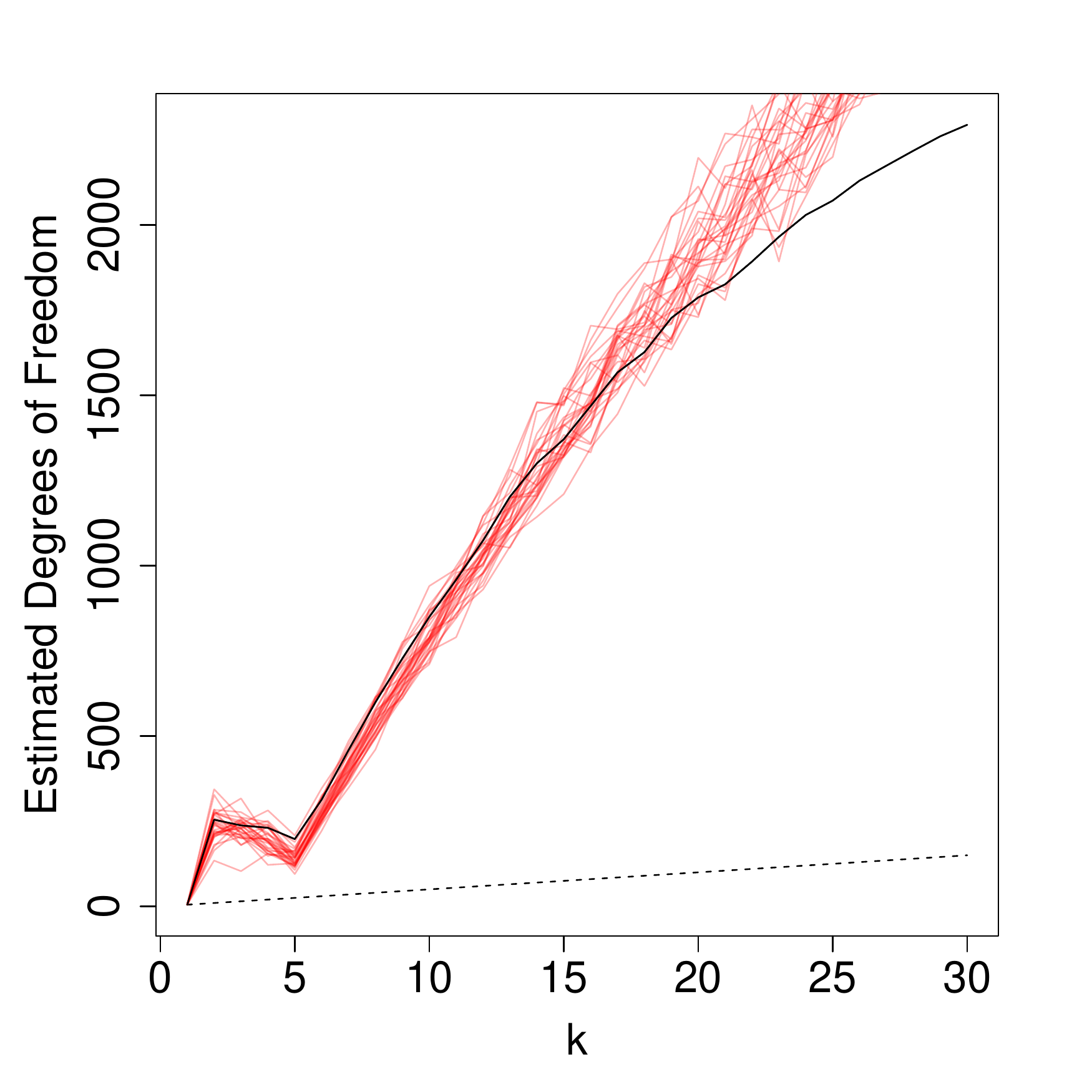}}
\subfigure[10 clusters in 5 dimensions]{\includegraphics[width = 3.9cm, height = 2.8cm]{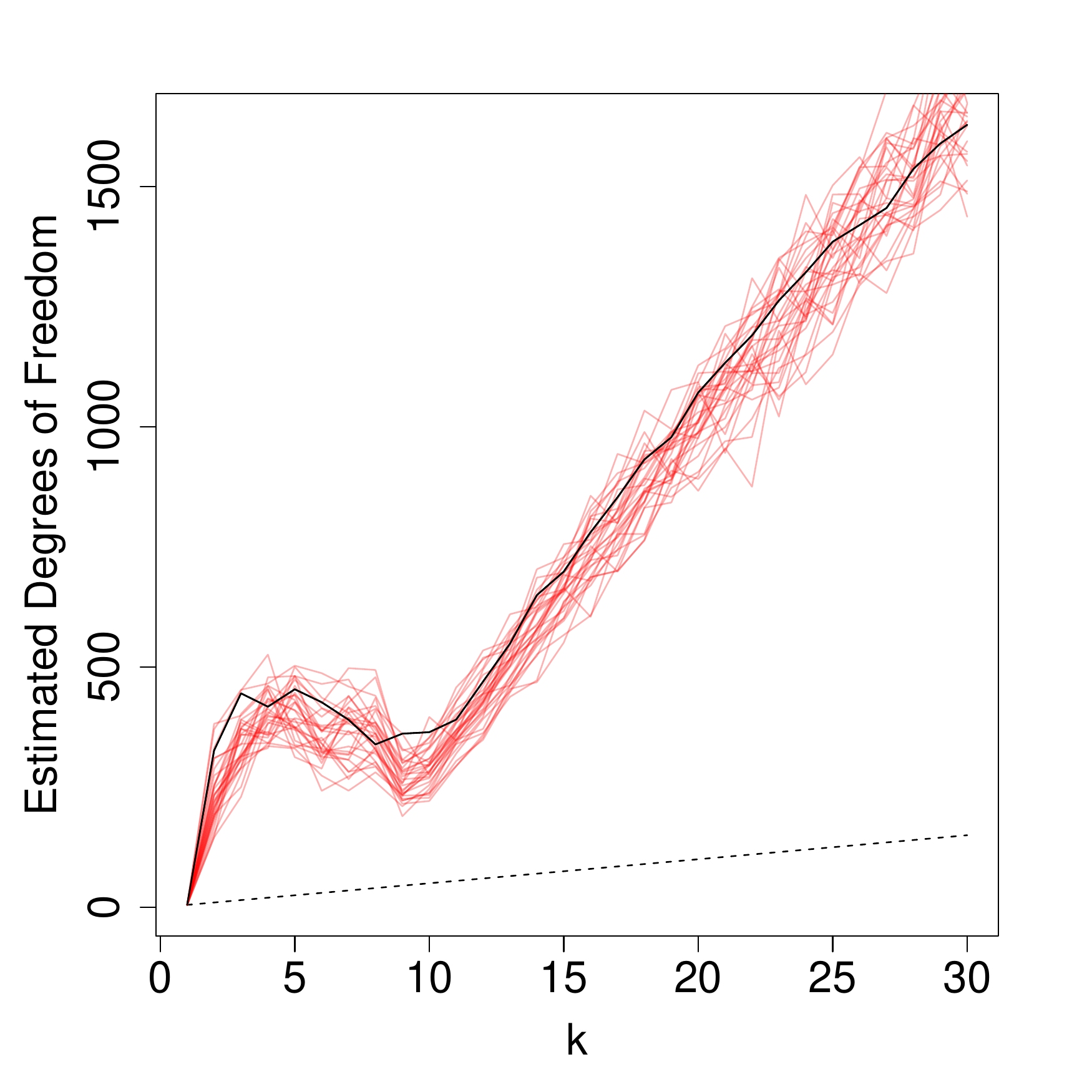}}
\subfigure[20 clusters in 5 dimensions]{\includegraphics[width = 3.9cm, height = 2.8cm]{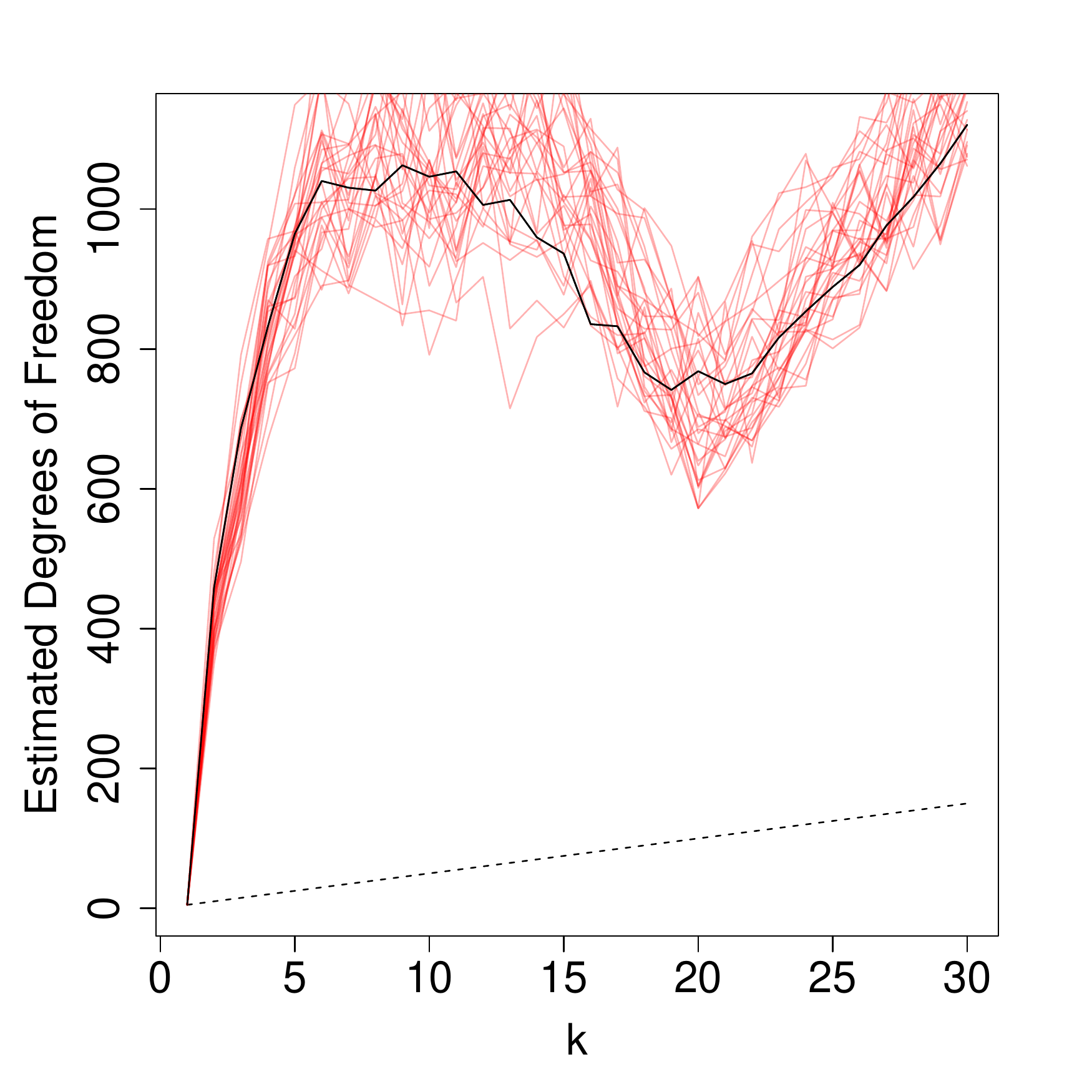}}\\
\subfigure[5 clusters in 10 dimensions]{\includegraphics[width = 3.9cm, height = 2.8cm]{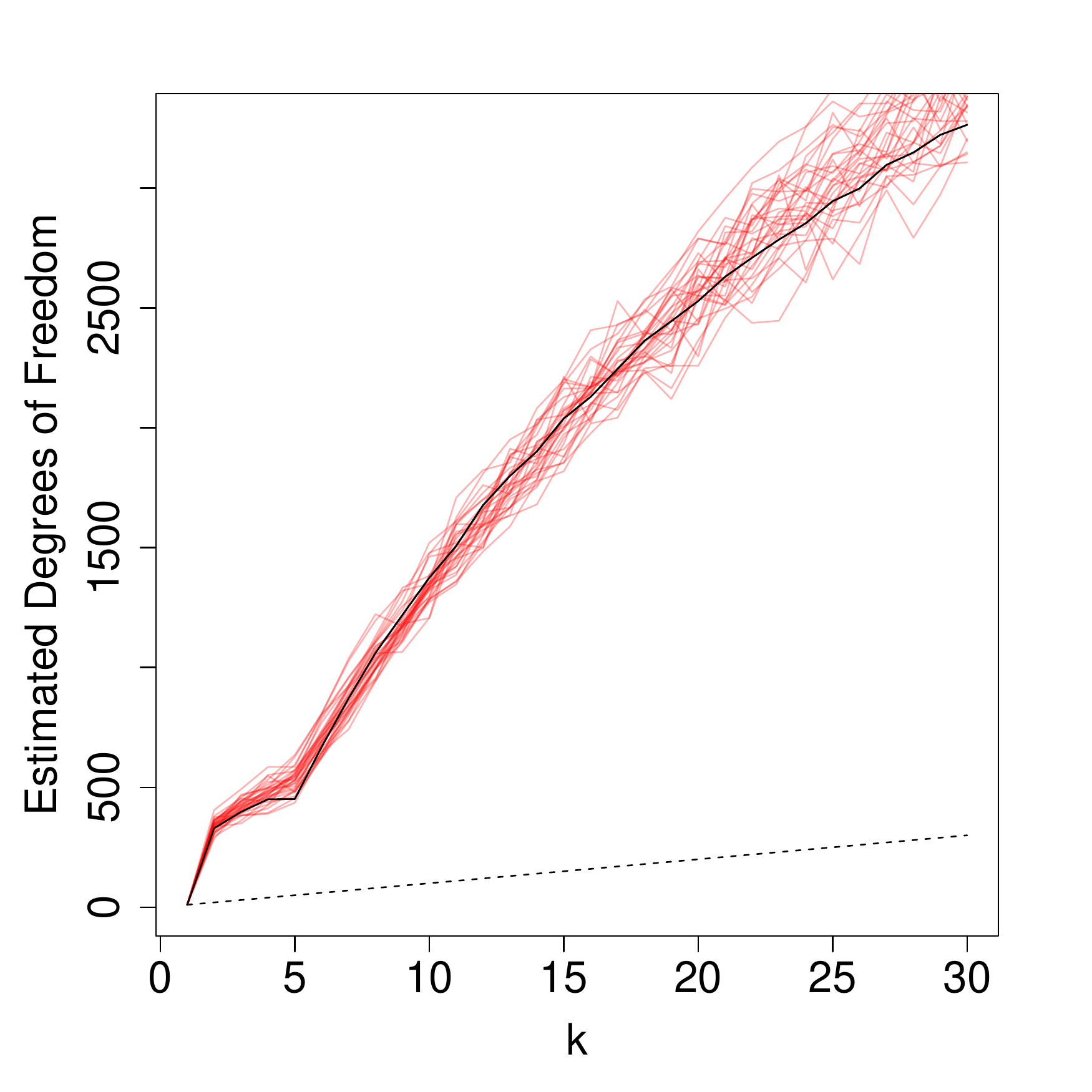}}
\subfigure[10 clusters in 10 dimensions]{\includegraphics[width = 3.9cm, height = 2.8cm]{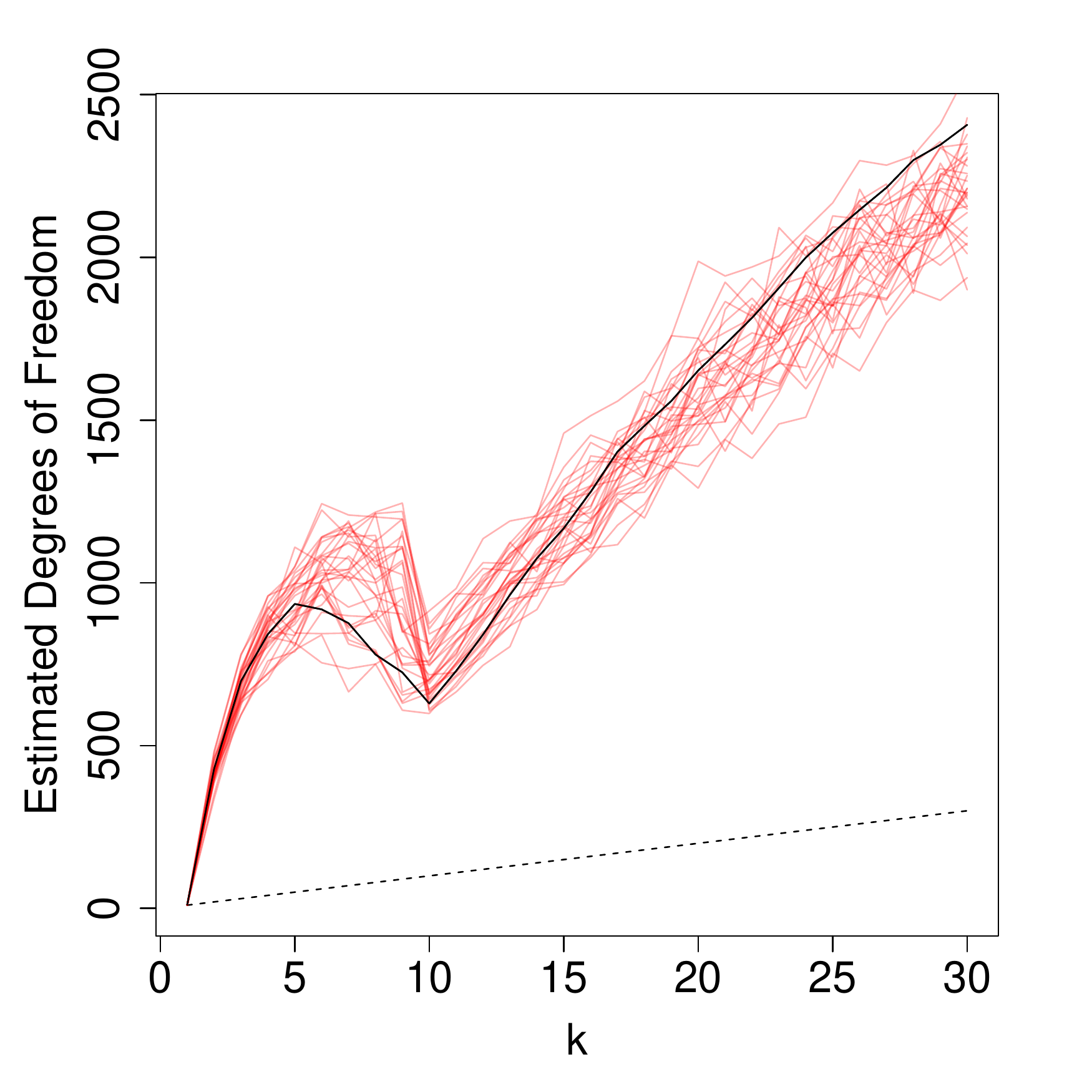}}
\subfigure[20 clusters in 10 dimensions]{\includegraphics[width = 3.9cm, height = 2.8cm]{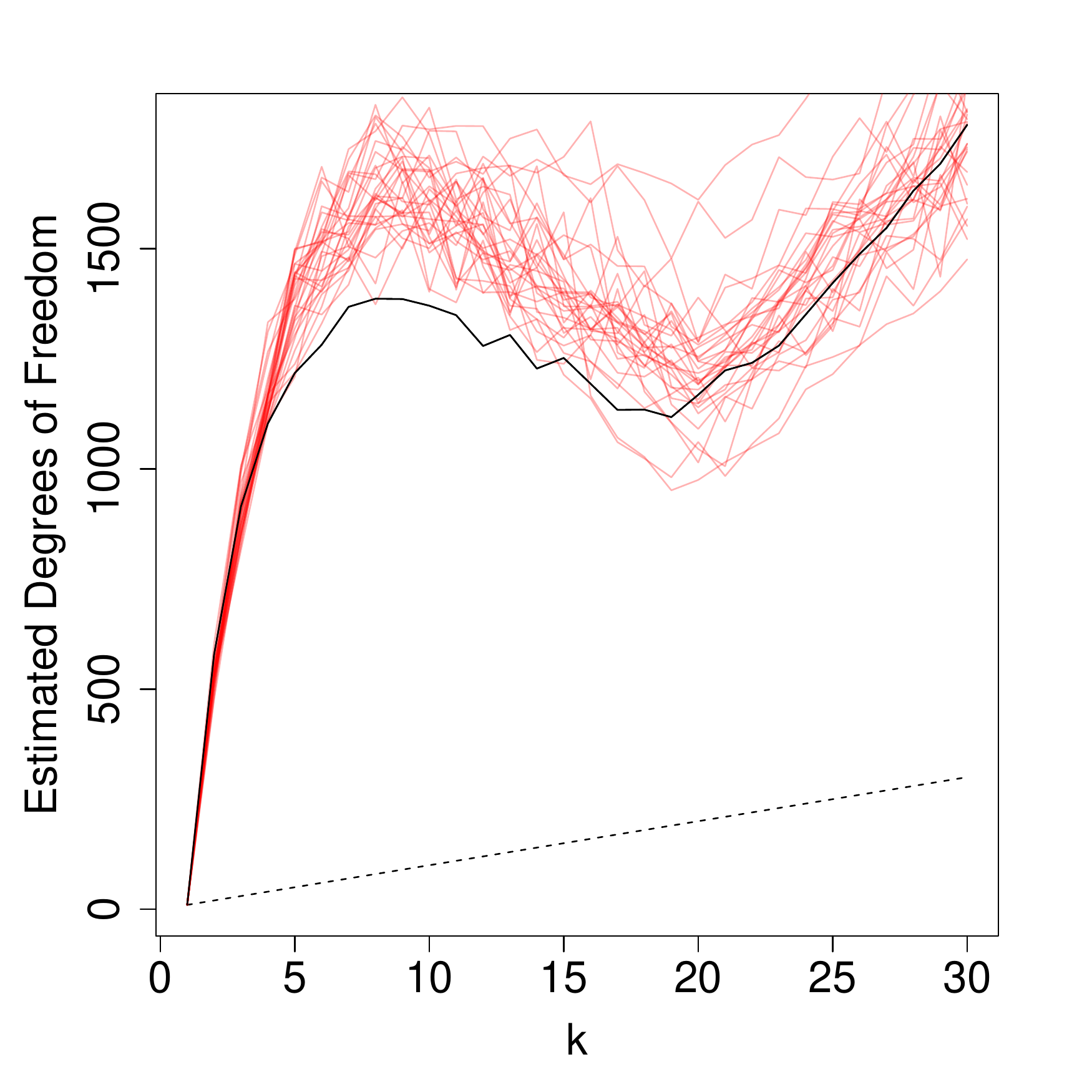}}\\
\subfigure[5 clusters in 20 dimensions]{\includegraphics[width = 3.9cm, height = 2.8cm]{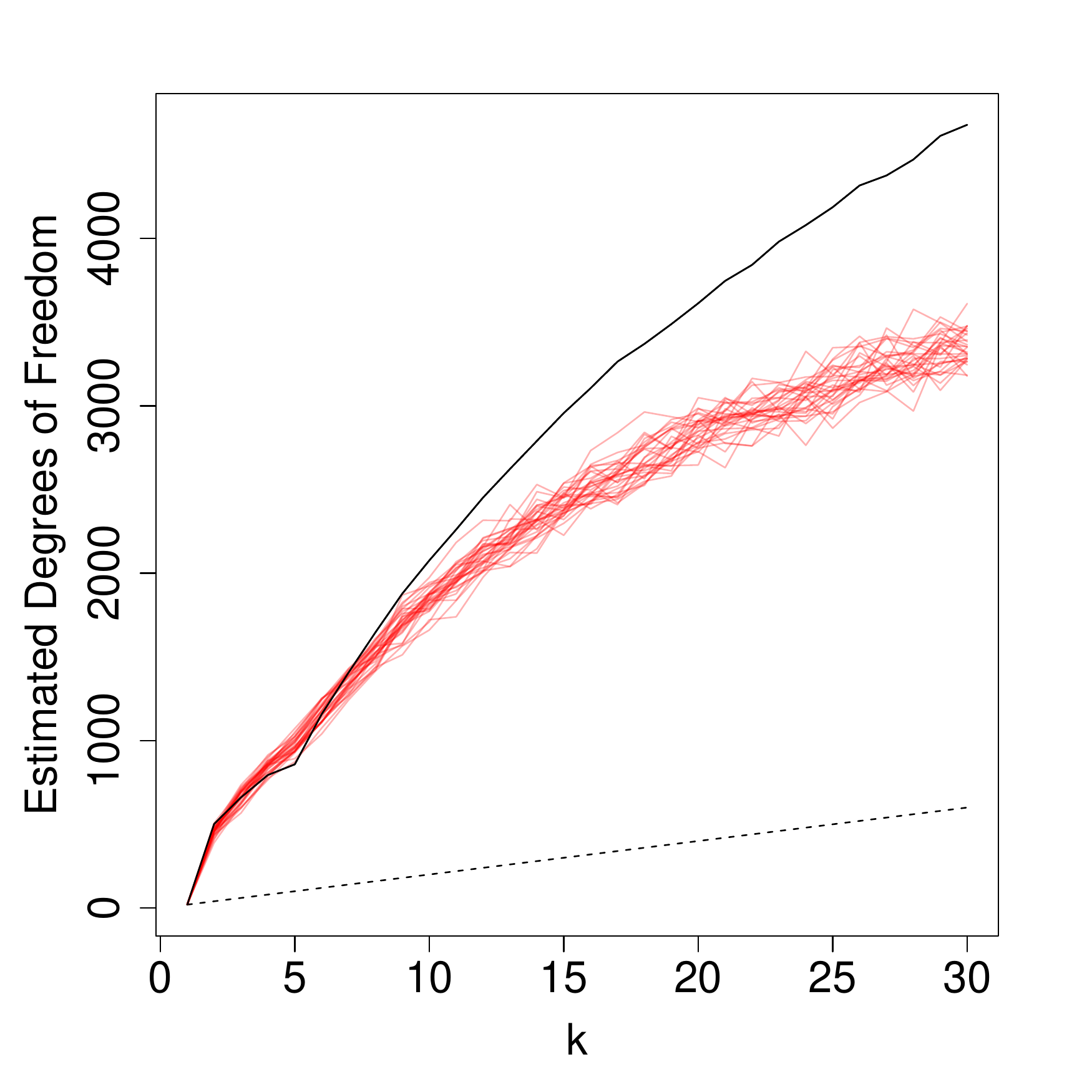}}
\subfigure[10 clusters in 20 dimensions]{\includegraphics[width = 3.9cm, height = 2.8cm]{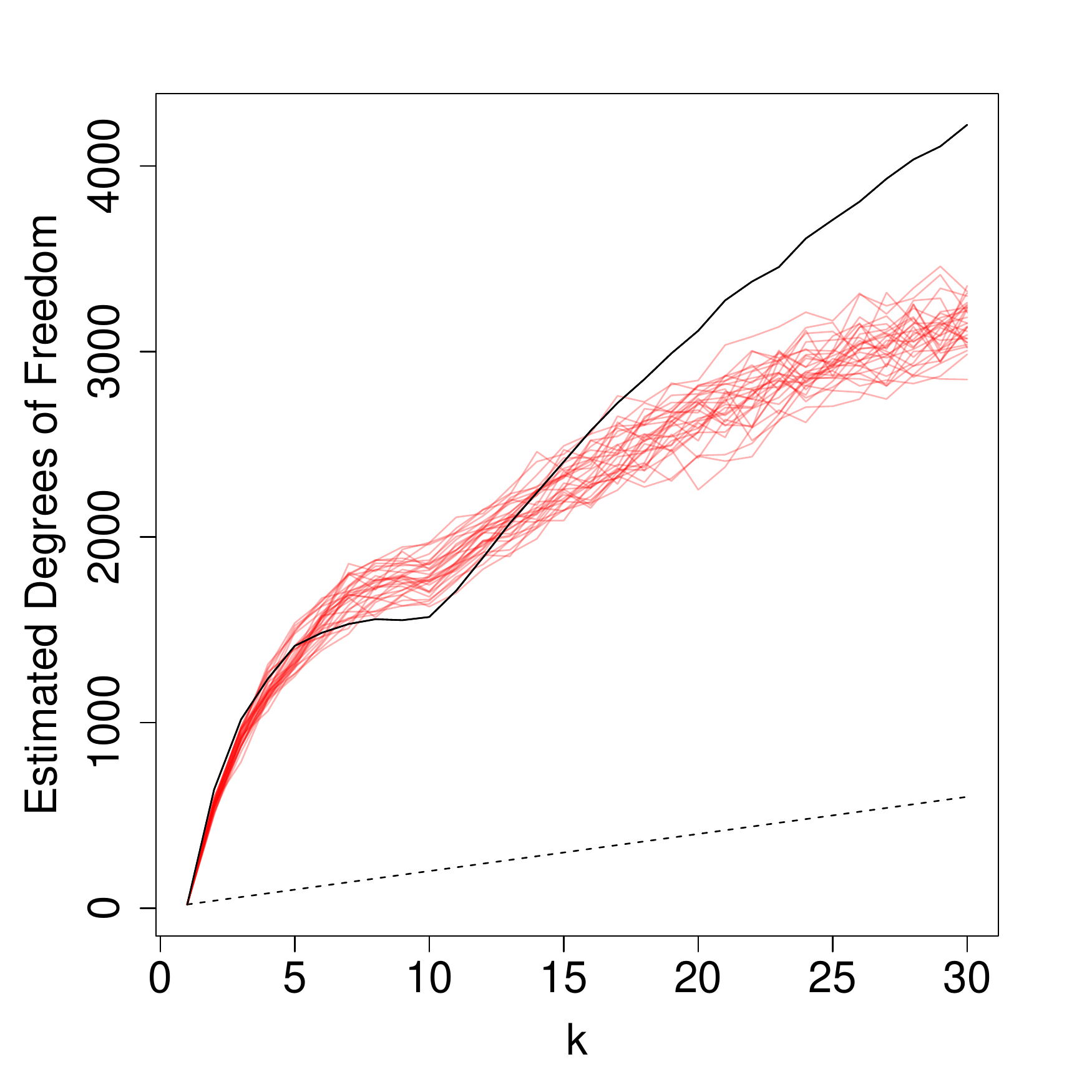}}
\subfigure[20 clusters in 20 dimensions]{\includegraphics[width = 3.9cm, height = 2.8cm]{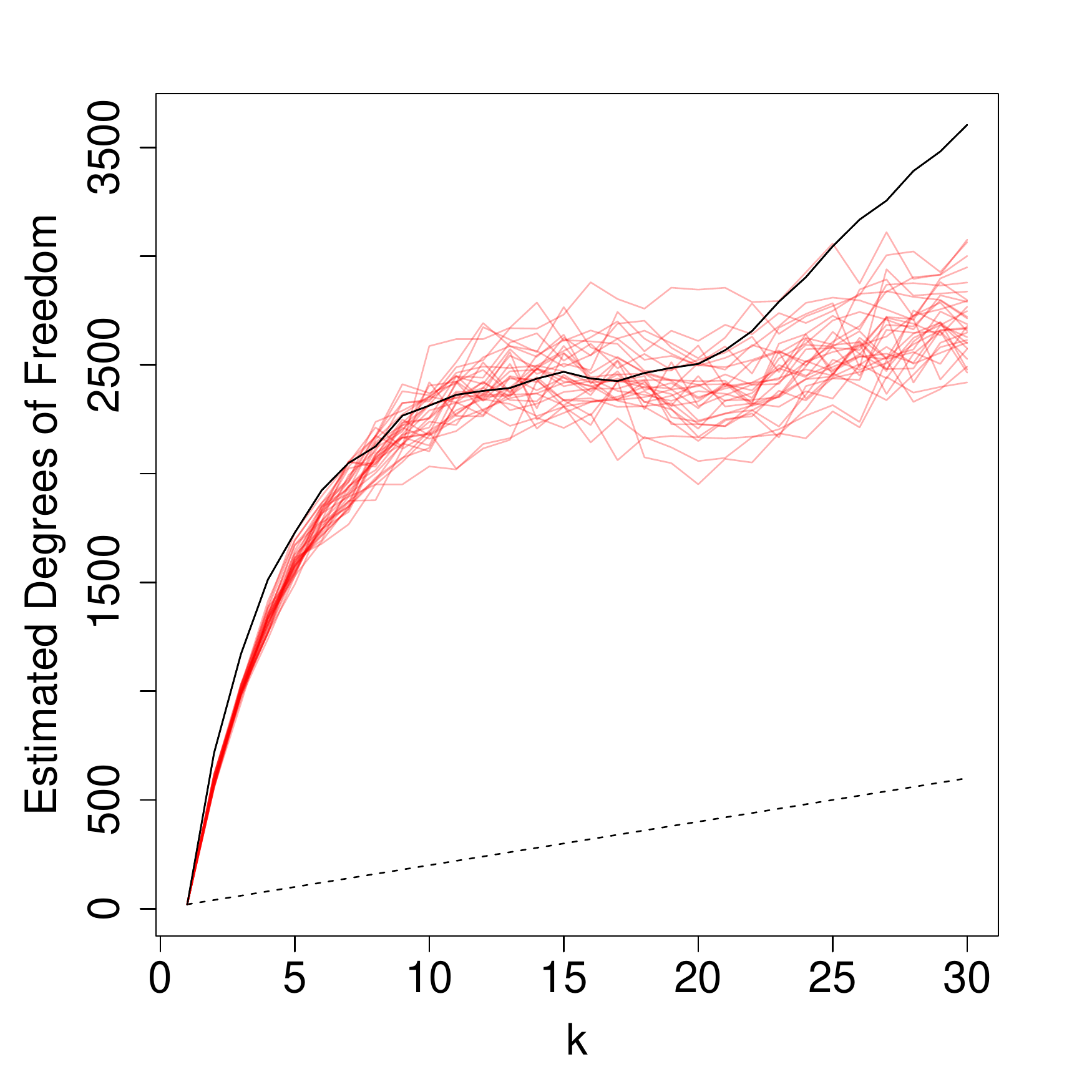}}\\
\caption{Estimated degrees of freedom computed through (i) direct sampling (------), (ii) proposed method for approximating effective degrees of freedom (\textcolor{red}{------}) and (iii) na{\"i}ve estimate of degrees of freedom (\textcolor{blue}{$\cdots\cdots$})\label{fig:ecov}}
\end{figure}

Figure~\ref{fig:ecov} shows the results of our simulation study. Data sets of size 1000 were generated under the modelling assumptions in Eq~(\ref{eq:generative}). The number of clusters and dimensions were each set to 5, 10 and 20. The figure shows plots of $k$ against the estimated degrees of freedom based on the above approach, where $k'$ was set to $k_{max}+1 = 31$. The results from 30 replications are shown (\textcolor{red}{------}). The plots also show direct empirical estimates of the degrees of freedom obtained by estimating the covariance between the model and the data when sampling from the true distribution (--------). That is, we generate multiple data sets according to Eq~(\ref{eq:generative}), apply $k$-means for each value of $k$, and compute the corresponding empirical covariance. To compute the direct estimate of degrees of freedom, this covariance is then simply divided by the true value $\sigma^2$. This direct estimate may therefore be seen as our target. For context we also include the plot of $kd$ ($\cdots\cdots$), corresponding to the na{\"i}ve degrees of freedom equated with the explicit model dimension.

Given the number of simplifications made, and the difficulty of the problem in the abstract, we find the estimation to be very satisfactory in general. The only exceptions apparent from this simple simulation study arose from the 20 dimensional examples, where the proposed method appears to underestimate the degrees of freedom for values of $k$ greater than the true value.
Note that from the point of view of model selection, a relatively larger underestimation of the degrees of freedom for a specific value of $k$ will bias the model selection towards that value of $k$. It is therefore this apparent negative bias in the estimated degrees of freedom for higher dimensional cases and for values of $k$ greater than the correct value which we find to be most problematic. We discuss a simple heuristic implemented to mitigate this effect in the next subsection, where we summarise our approach for performing model selection using the estimated degrees of freedom.

%\subsection{A Brief Comment on Computational Complexity}

\section{Choosing $k$ Using the BIC} \label{sec:BIC}

The Bayesian Information Criterion approximates, up to unnecessary constants, the logarithm of the evidence for a model $\M$, i.e., $P(\M|\X)$, using
\begin{align*}
-2\ell(\X|\M) + \log(m)\mathrm{df}(\M).
\end{align*}
Again $\ell(\X|\M)$ is the model log-likelihood and here $m$ is the number of independent ``residuals'' in $\X$. 
With the modelling assumptions in Eq~(\ref{eq:generative}), the BIC for $k$-means is therefore, up to an additive constant,
\begin{align*}
\frac{1}{\sigma^2}\sum_{i=1}^n\sum_{j=1}^d (\X_{i,j}-\mmu_{i,j})^2 + nd\log(\sigma^2) + \log(nd) \mathrm{df}(\M).
\end{align*}
% bias term is given as $\log(nd)\mathrm{df}(\M)$ since the number of independent residuals in the matrix $\E$ is $n\times d$.
Setting $\hat\mmu$ here to be equal to $\M(\X; k)$, i.e., the matrix of fitted values from the model, and $\hat \sigma^2 = \frac{1}{nd}\sum_{i=1}^n\sum_{j=1}^d(\X_{i,j} - \hat \mmu_{i,j})^2$ to be the corresponding maximum likelihood estimate of the in-cluster variance, the estimated BIC in the $k$-means model with $k$ clusters is therefore, up to an additive constant,
\begin{align*}
nd \log\left(\sum_{i=1}^n\sum_{j=1}^d (\X_{i,j}-\hat\mmu_{i,j})^2\right) + \log(nd)\widehat{\mathrm{df}(\M(\X; k))}.
\end{align*}
Now, we found in the previous section that the proposed approximation method for the model degrees of freedom has the potential to exhibit negative bias for larger values of $d$ and for $k$ greater than the true number of clusters. To mitigate the effect this has on model selection, we select the number of clusters as the smallest value of $k$ which corresponds to a local minimum in the estimated BIC curve, seen as a function of $k$. If no such local minima are present, then we select either $k_{\mathrm{min}}$ or $k_{\mathrm{max}}$; whichever gives the lowest value of the BIC. A similar ``first extremum'' approach for model selection has also been used by~\cite{tibshirani2001estimating}. We also apply a simple local-linear smoothing to the approximated degrees of freedom curves. This mitigates the effect of variation, which is quite pronounced in, e.g., Figure~\ref{fig:ecov} (c) and (f). It also smooths over the short range variation within each estimated curve which is apparent in the proposed estimates, but not present in the curves estimated by direct sampling. Not smoothing over this variation has the potential to induce spurious local minima in the resulting BIC curves, which would not be present were it possible to obtain such direct estimates in practice.

%To account for this  We also smooth the estimated degrees of freedom curves to mitigate the effect of variation which is quite pronounced in, for example, Figure~\ref{fig:ecov} (c), (f) and (i). Any simple smoothing is appropriate, where we apply a local linear kernel smoother with a Gaussian kernel. 
%This bandwidth tends to over-smooth when compared with selection methods which attempt to minimise squared error (either through cross validation, or based on asymptotics), however we have found it to be fairly reliable in that it smooths over relatively large, and apparently spurious fluctuations very well, but maintains the location of the minima reasonably well.
%This has been found to provide reliable results in most cases, but expect conscientious users will investigate the BIC curves and also consider alternative values of $k$ which correspond with repeated minima.

\section{Experimental Results} \label{sec:experiments}

In this section we report on the results from experiments conducted to assess the performance of the proposed approach for model selection, using both simulated data and data from real applications. In addition to the proposed approach, we also experimented with the following popular existing methods for model selection:

\begin{enumerate}

\item The Gap Statistic~\citep{tibshirani2001estimating}, which is based on approximating, through Monte Carlo simulation, the deviation of the (transformed) within cluster sum of squares from its expected value when the underlying data distribution contains no clusters.
%The implementation in the {\tt R} package {\tt cluster}~\citep{clusterpackage} was used, after a minor modification which allowed the number of initialisations for $k$-means to differ between the true data set and the Monte Carlo samples. 
Due to high computation time, solutions for the Monte Carlo samples were based on a single initialisation. Using ten initialisations, as for the clustering solutions of the actual data sets, did not produce better results in general on data sets for which this approach terminated in a reasonable amount of time.\\
\item The method of~\cite{pham2005selection} which uses the same motivation as the Gap Statistic, but determines the deviation of the sum of squares from its expected value analytically under the assumption that the data distribution meets the standard $k$-means assumptions. We use ``fK'' to refer to this method in the remainder.\\ %When the data meet these assumptions the expectation is that this approach should perform extremely well. This approach will be referred to as fK.
\item The Silhouette Index~\citep{kaufman2009finding}, which is based on comparing the average dissimilarity of each point to its own cluster with its average dissimilarity to points in different clusters. Dissimilarity is determined by the Euclidean distance between points.\\
\item The Jump Statistic~\citep{Sugar}, which selects the number of clusters based on the first differences in the $k$-means objective raised to the power $-\frac{d}{2}$. This statistic is based on rate distortion theory, which approximates the mutual information between the complete data set and the summarisation by the $k$ centroids.\\
\item The Bayesian Information Criterion with a na{\"i}ve estimate of the degrees of freedom given by $kd$. We used exactly the same selection approach as for the proposed method.
\end{enumerate}

The clustering solutions given to each model selection method were the best, in terms of $k$-means objective, from ten random initialisations for each value of $k$\footnote{Exactly the same clustering solutions were given to all selection methods.}. For all data sets values of $k$ from 1 to 30 were considered. In all cases clustering solutions were obtained using the implementation of $k$-means provided in {\tt R}'s base {\tt stats} package~\citep{R}. 

\subsection{Simulations}

\begin{figure}
\centering
\subfigure[Assumptions met]{\includegraphics[width=3.9cm]{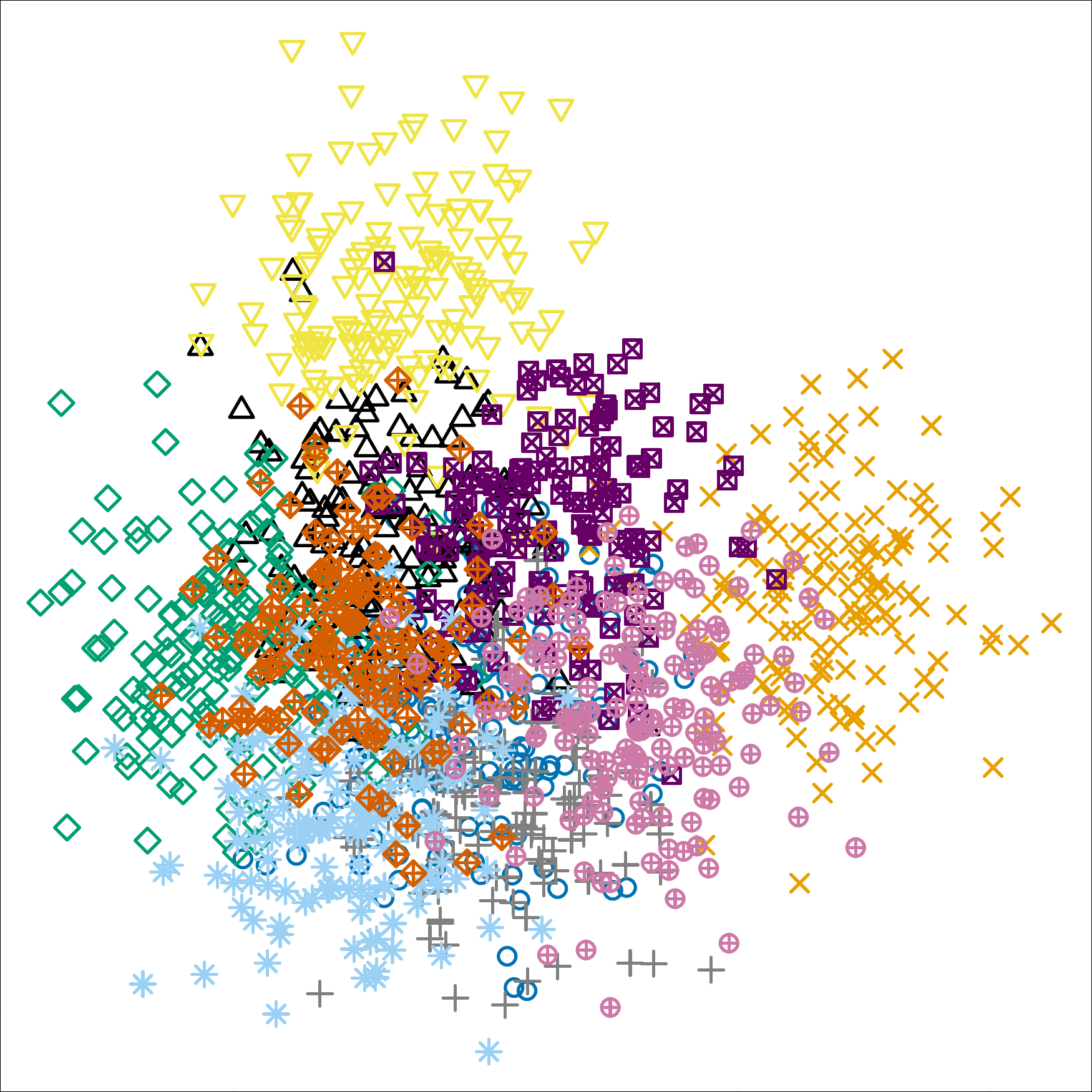}}
\subfigure[Varying cluster scale]{\includegraphics[width=3.9cm]{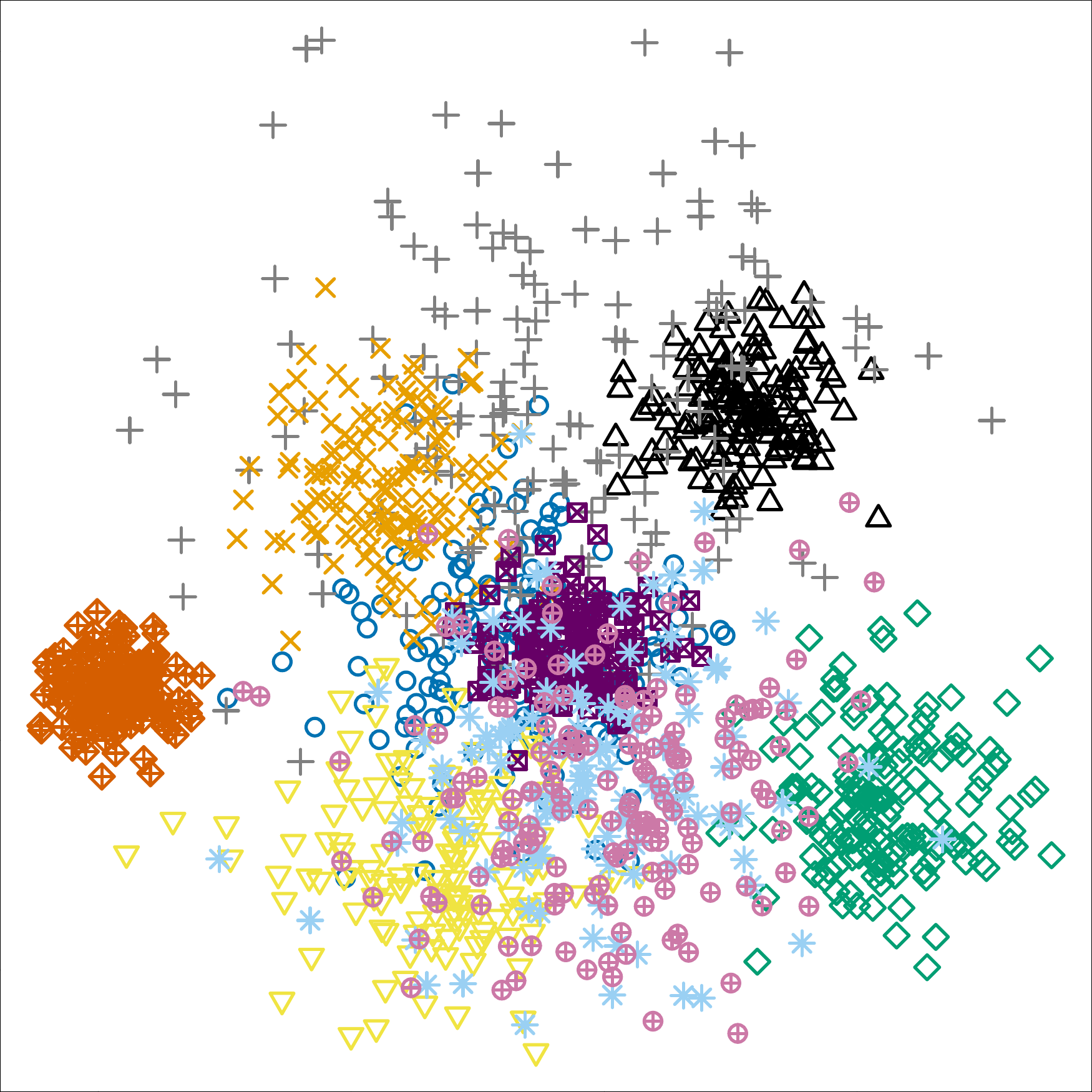}}
\subfigure[Varying cluster shape]{\includegraphics[width=3.9cm]{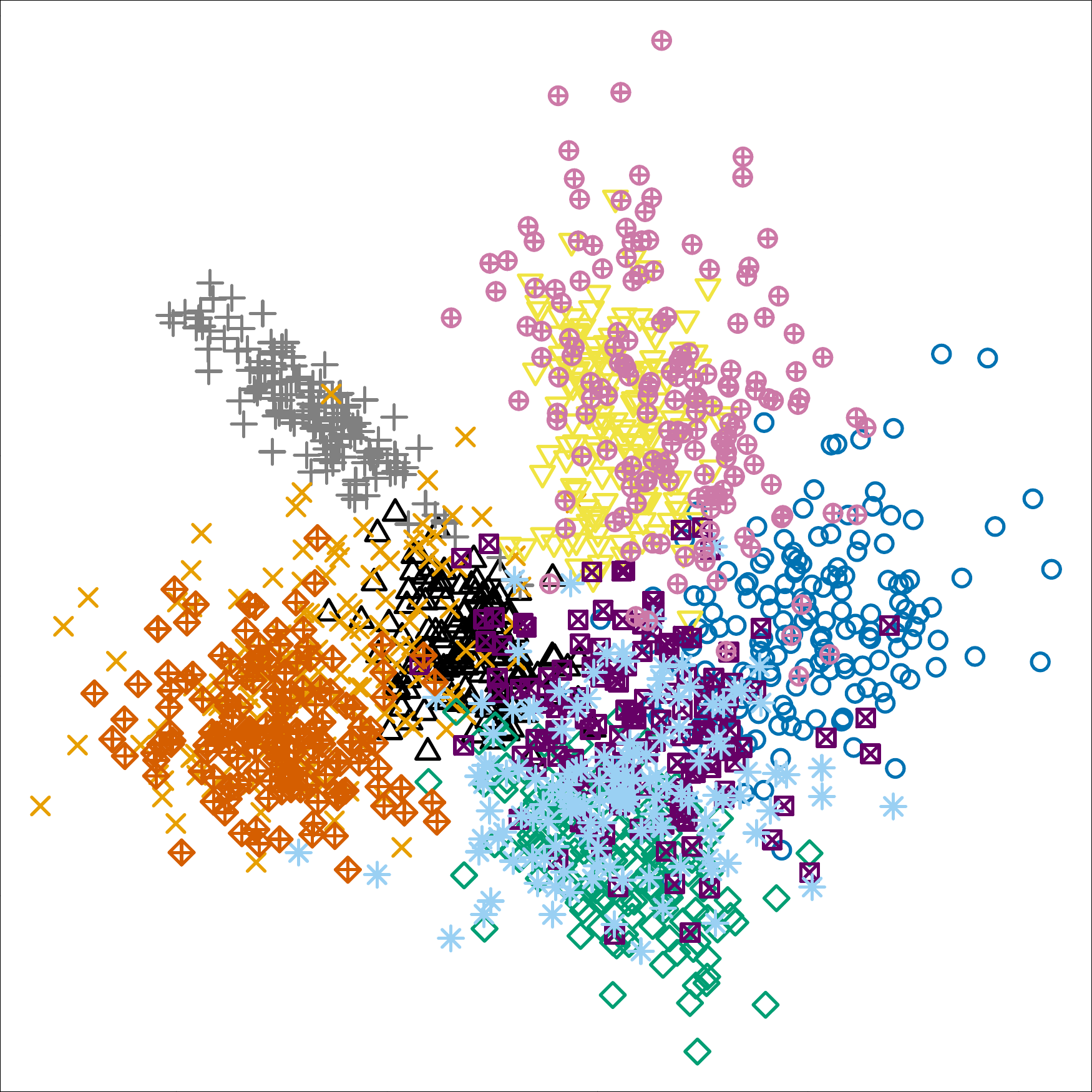}}
\caption{Plots of typical simulated data sets from Gaussian mixtures with 10 clusters in 10 dimensions. Plots show data projected onto their first two principal components. Clusters are differentiated by colour and by point character \label{fig:Gauss_plots}}
\end{figure}

\begin{figure}
\centering
\subfigure[Long tails ($t_3$)]{\includegraphics[width=3.9cm]{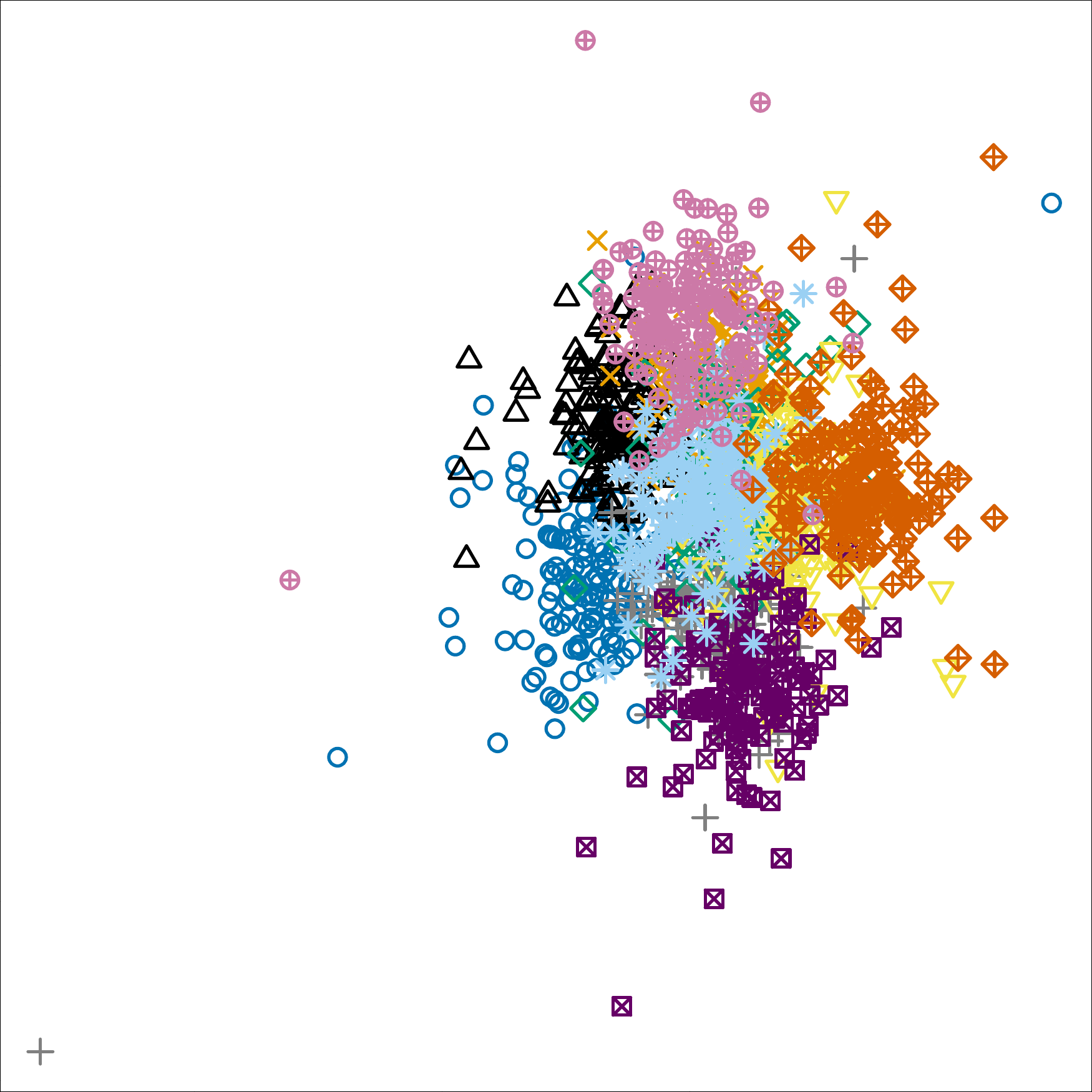}}
\subfigure[Uniform clusters]{\includegraphics[width=3.9cm]{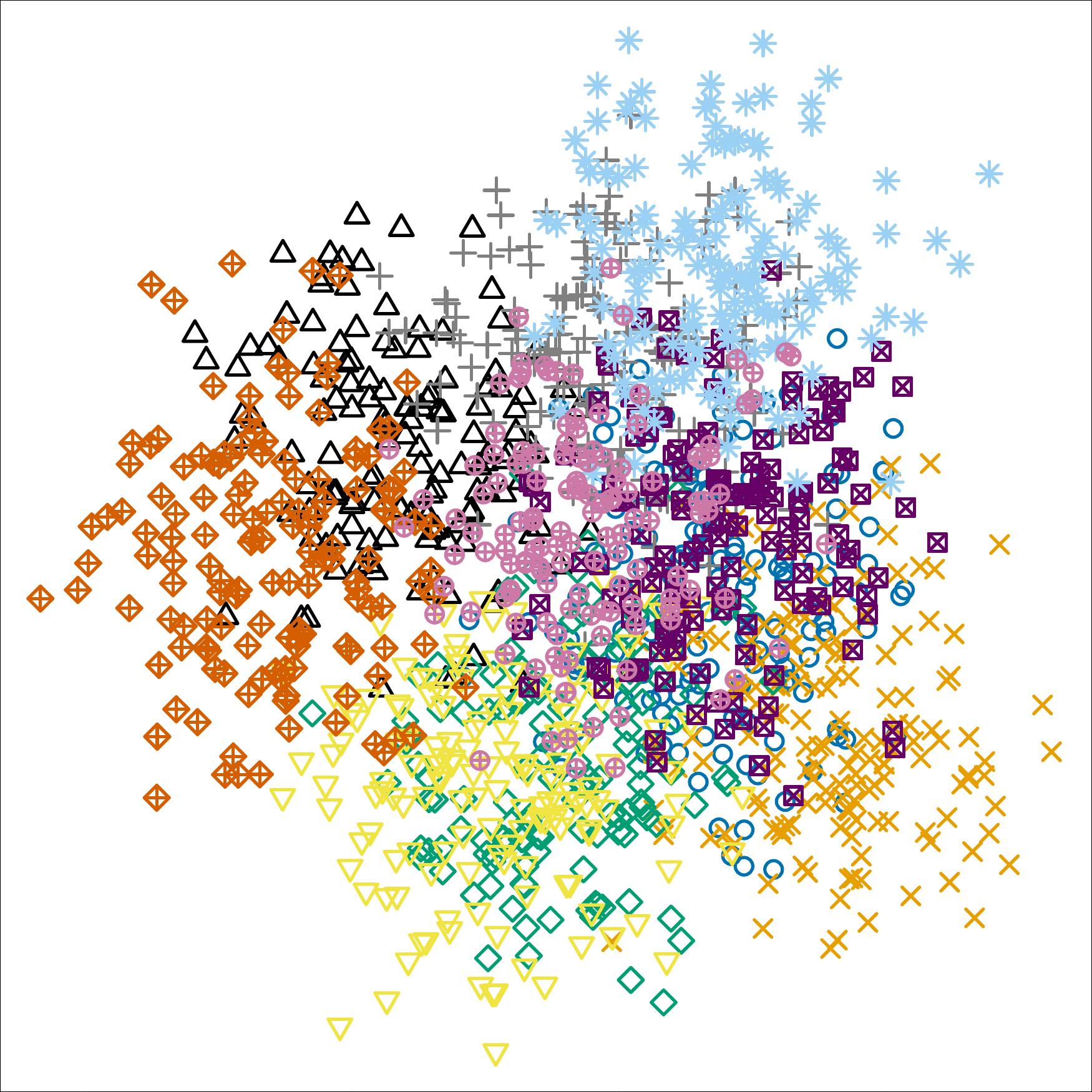}}
\subfigure[Non-convex clusters]{\includegraphics[width=3.9cm]{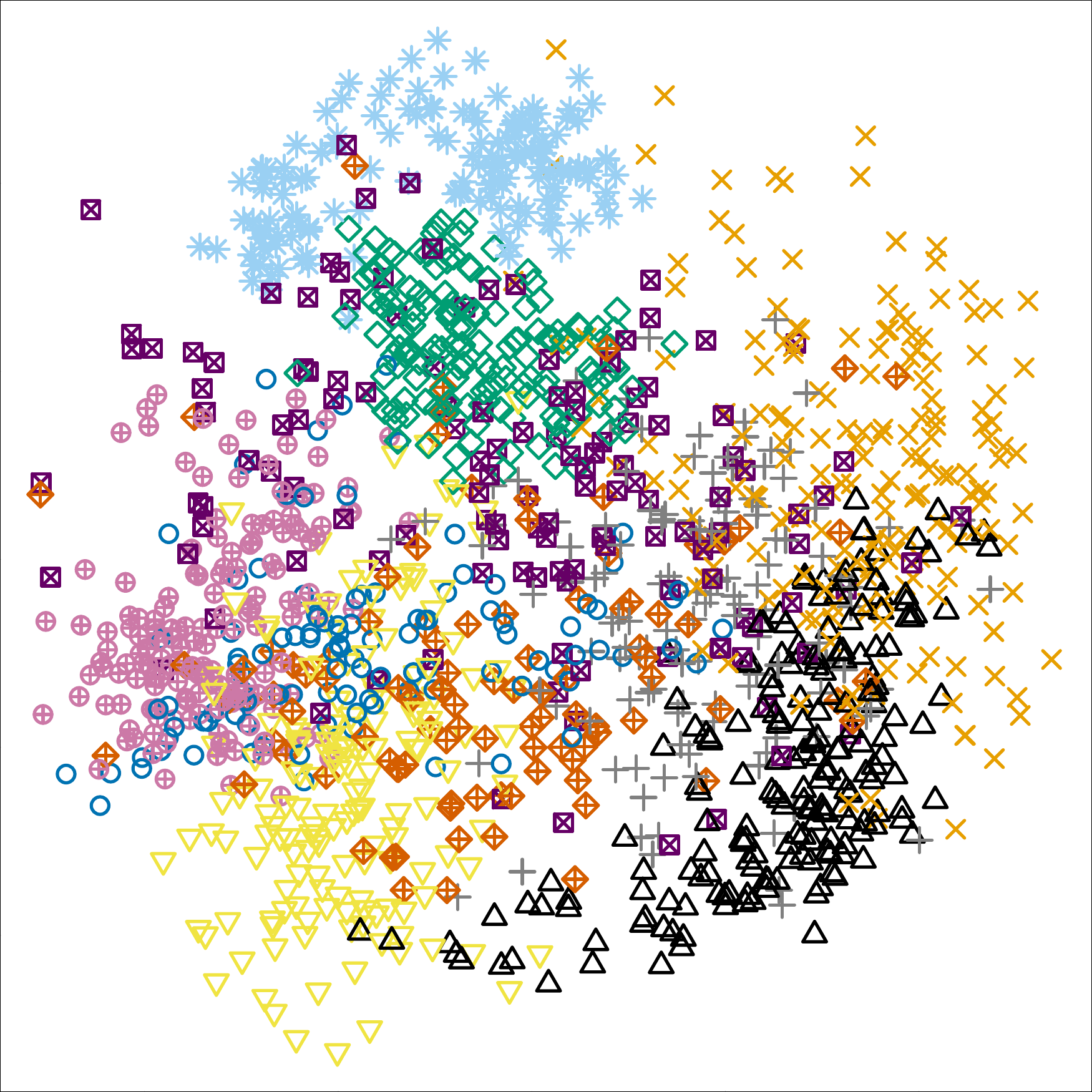}}
\caption{Plots of typical simulated data sets from non-Gaussian mixtures with 10 clusters in 10 dimensions. Plots show data projected onto their first two principal components. Clusters are differentiated by colour and by point character \label{fig:non-Gauss_plots}}
\end{figure}

In this section we report results from simulated data sets where the model structure is known and can be reasonably well controlled. We investigate scenarios including (i) when the $k$-means model assumptions of a Gaussian mixture with equal mixing proportions and equal and spherical covariance matrices are met; (ii) simple deviations from these assumptions including Gaussian mixtures with non-spherical covariances and unequal scale/mixture component density; and (iii) deviations from Gaussianity including slightly non-convex clusters and different tails in the residual distributions. To generate non-convex clusters, we use the approach described in~\cite{hofmeyr2019improving}, and using the {\tt R} package {\tt spuds}\footnote{\url{https://github.com/DavidHofmeyr/spuds}}. Here, points generated from a Gaussian mixture are given perturbations, and the size of the perturbation is greater the nearer a point is to points from other clusters. This simulation scheme was designed to test more flexible clustering methods, such as spectral clustering. However, $k$-means is capable of achieving high clustering accuracy when the degree of non-convexity of the clusters is not too substantial.
For the reader's interest, Figures~\ref{fig:Gauss_plots} and~\ref{fig:non-Gauss_plots} show typical data sets generated from each simulation scheme, for the cases with 10 clusters in 10 dimensions. The figures show the two-dimensional principal component plots of the data. These give some indication of the types of data given to the algorithms, and can be used to infer somewhat the comparative difficulty of the clustering problems.

The results from the simulations are summarised in Tables~\ref{tb:gauss_sims} and~\ref{tb:non_gauss_sims}. For each simulation scheme 30 data sets were generated, and the best performing methods, in terms of quality of solutions selected (see below), are indicated by bold font. In addition, methods whose performance was not significantly different from the best, based on a paired Wilcoxon signed rank test using a $p$-value threshold of 0.01, are also highlighted. We chose to use a small $p$-value to retain considerable discrimination in the results among the methods which perform well in general, but not so small that a single or few instances of one method identifying a single extra cluster would lead to it being excluded from the ``best performers'' for a given simulation scenario. % \footnote{Naturally any $p$-value threshold is arbitrary, but this allows for at least some degree of statistical confidence in the conclusions drawn from the results.}, are also highlighted.
 Methods are compared based on their ability to select the correct number of clusters, and also based on the quality of the clustering solutions selected when compared to the ground truth\footnote{The ground truth here corresponds to the identities of the mixture components from which the data were generated.}. 
 %
 %
% For this we use the adjusted Rand index~\citep[ARI]{hubert1985comparing} which is given as the proportion of pairs of points which are either grouped together or assigned to different clusters in both the clustering solution and under the ground truth, normalised based on the expected proportion of such pairs based on a random assignment. 
 % 
 For this we use the adjusted Rand index~\citep[ARI]{hubert1985comparing}. The Rand index~\citep{rand1971objective} is given as the proportion of pairs of points which are either grouped together in both the clustering solution and the ground truth or assigned to different clusters both in the solution and the ground truth. An adjustment is then applied to normalise this proportion based on its expectation under a random assignment.
 The clustering accuracy is important to consider since it provides a means for comparing solutions when incorrect values of $k$ are selected. 
%Furthermore, when the $k$-means model assumptions are not met, it is possible that the best $k$-means solution from the point of view of accuracy does not contain the same number of clusters as the ground truth. This becomes even more important when considering data from real applications, where the data frequently deviate very substantially from these assumptions. 
Table~\ref{tb:gauss_sims} shows the results corresponding to data sets generated from Gaussian mixtures. Both the proposed approach, described as BIC$_{\widehat{edf}}$, and the Silhouette Index show very strong performance.  The Jump Statistic performs very well when the assumptions are met exactly, but the performance drops dramatically when these assumptions are deviated from. It is worth noting that of the methods compared, the Silhouette Index is the only approach which is incapable of discerning ``one cluster'' from ``more than one cluster''. It is possible, therefore, that the performance of this method is in some sense slightly over-estimated, since its fail-cases are not as severe.

Table~\ref{tb:non_gauss_sims} shows the results corresponding to data sets generated from non-Gaussian mixtures. In this case the Silhouette Index enjoys the best performance. The performance of the proposed method is also strong, but significantly below that of the Silhouette in a number of cases, most frequently on the data containing non-convex clusters. The Gap Statistic here showed numerous instances of a failure to identify the presence of clusters. This is an interesting point to note, as in our experiments on data from real applications, the Gap Statistic performs well in general on non-Gaussian data.

\setlength{\tabcolsep}{3pt}

\begin{table*}
\centering
\caption{Results from simulated Gaussian mixture data sets. The Median of the number of clusters selected by each method ($\hat k$) and corresponding adjusted Rand index (ARI) are reported. Subscripts show the 10$^{th}$ and 90$^{th}$ centiles. The quantiles are based on the results from 30 data sets generated for each simulation set-up. Highest performances for each scenario are highlighted , as are those which are not significantly different from the highest based on a paired Wilcoxon signed rank test with $p$-value threshold of 0.01.
\label{tb:gauss_sims}
}
\scalebox{0.7}{
\begin{tabular}{lcccccccccccccc}
&&& \mcl{2}{fK} & \mcl{2}{Gap} & \mcl{2}{Silh.} & \mcl{2}{Jump} &  \mcl{2}{BIC} & \mcl{2}{BIC$_{\widehat{edf}}$}   \\
Simulation & k & d & $\hat{k}$ & ARI & $\hat{k}$ & ARI  & $\hat{k}$ & ARI& $\hat{k}$ & ARI   & $\hat{k}$ & ARI  & $\hat{k}$ & ARI   \\
\hline
Assump-  & 5 & 5 &  $5_{3,5}$ & $98_{59,99}$ &  $5_{5,5}$ & ${\bf 98_{97,99}}$ &  $5_{5,5}$ & ${\bf 98_{97,99}}$ &  $5_{5,5}$ & ${\bf 98_{97,99}}$ &  $24_{18,29}$ & $30_{25,38}$ &  $5_{5,5}$ & ${\bf 98_{98,99}}$  \\
tions & & 10 &  $4_{2,5}$ & $74_{31,96}$ &  $5_{5,5}$ & ${\bf 94_{92,97}}$ &  $5_{5,5}$ & ${\bf 94_{92,97}}$ &  $5_{5,5}$ & ${\bf 94_{92,97}}$ &  $19_{15,21}$ & $35_{32,43}$ &  $5_{5,5}$ & ${\bf 94_{92,97}}$  \\
met & & 15 &  $2_{2,4}$ & $29_{20,70}$ &  $5_{1,5}$ & ${\bf 85_{0,87}}$ &  $5_{5,5}$ & ${\bf 86_{83,87}}$ &  $5_{5,29}$ & ${\bf 86_{21,87}}$ &  $12_{9,14}$ & $ 45_{39,58}$ &  $5_{1,5}$ & $ 84_{0,87}$  \\
& 10 & 5 &  $ 8_{2,10}$&$ 79_{17,97}$&$ 10_{1,10}$&$ 94_{0,98}$&$ 10_{10,10}$&${\bf 96_{94,98}}$&$ 10_{10,10}$&${\bf 96_{94,98}}$&$ 20_{17,26}$&$ 66_{56,74}$&$ 10_{10,11}$&${\bf 96_{92,98}}$ \\
& & 10 & $ 7_{2,10}$&$ 62_{14,94}$&$ 6_{1,10}$&$ 44_{0,93}$&$ 10_{10,10}$&${\bf 92_{90,94}}$&$ 10_{10,10}$&${\bf 92_{90,94}}$&$ 16_{14,17}$&$ 73_{68,81}$&$ 10_{10,10}$&${\bf 92_{89,94}}$\\
& & 15 & $ 2_{2,2}$&$ 12_{11,14}$&$ 1_{1,10}$&$ 0_{0,81}$&$ 10_{10,10}$&${\bf 82_{80,85}}$&$ 10_{10,10}$&${\bf 82_{80,85}}$&$ 11_{10,12}$&$ 78_{74,82}$&$ 10_{10,10}$&${\bf 82_{80,85}}$\\
& 15 & 5 & $ 12_{2,14}$&$ 78_{12,90}$&$ 2_{1,15}$&$ 12_{0,95}$&$ 15_{14,15}$&${\bf 93_{88,96}}$&$ 15_{14,15}$&${\bf 93_{88,96}}$&$ 24_{20,29}$&$ 75_{66,84}$&$ 15_{15,16}$&${\bf 94_{89,96}}$\\
& & 10 & $ 11_{2,15}$&$ 66_{9,91}$&$ 1_{1,15}$&$ 0_{0,91}$&$ 15_{15,15}$&${\bf 90_{85,92}}$&$ 15_{15,15}$&${\bf 90_{86,92}}$&$ 17_{16,18}$&$ 86_{82,88}$&$ 15_{15,16}$&$ {\bf 89_{86,92}}$\\
& & 15 & $ 2_{2,12}$&$ 8_{7,63}$&$ 1_{1,1}$&$ 0_{0,0}$&$ 15_{15,15}$&${\bf 79_{76,81}}$&$ 15_{15,15}$&${\bf 79_{75,81}}$&$ 15_{15,15}$&${\bf 79_{76,81}}$&$ 16_{15,16}$&$ 78_{75,80}$\\
\hline
Within & 5 & 5 & $ 5_{2,5}$&$ {\bf 94_{38,99}}$&$ 5_{5,6}$&$ 94_{78,99}$&$ 5_{5,5}$&${\bf 96_{88,99}}$&$ 5_{5,29}$&$ 94_{35,99}$&$ 13_{8,19}$&$ 69_{50,82}$&$ 5_{5,5}$&${\bf 96_{90,99}}$ \\
cluster &  & 10 & $ 4_{2,5}$&$ 73_{27,97}$&$ 5_{5,6}$&${\bf 91_{83,97}}$&$ 7_{5,13}$&$ 87_{75,97}$&$ 26_{5,29}$&$ 59_{39,97}$&$ 17_{11,23}$&$ 64_{52,75}$&$ 5_{5,6}$&${\bf 91_{85,97}}$\\
scale & & 15 & $ 2_{2,5}$&$ 31_{20,82}$&$ 5_{5,6}$&${\bf 84_{78,93}}$&$ 14_{5,28}$&$ 73_{59,94}$&$ 28_{5,30}$&$ 54_{45,94}$&$ 14_{11,17}$&$ 64_{58,76}$&$ 5_{5,6}$&${\bf 84_{78,94}}$\\
varies & 10 & 5 & $ 9_{5,10}$&$ 80_{45,95}$&$ 10_{1,10}$&$ 91_{0,98}$&$ 10_{10,14}$&${\bf 93_{86,98}}$&$ 10_{10,24}$&${\bf 93_{71,98}}$&$ 19_{15,25}$&$ 79_{71,86}$&$ 11_{10,12}$&$ 91_{82,96}$\\
& & 10 & $ 7_{2,10}$&$ 62_{14,92}$&$ 10_{1,11}$&${\bf 85_{0,92}}$&$ 14_{10,24}$&$ 85_{76,93}$&$ 10_{10,29}$&${\bf 86_{68,93}}$&$ 19_{13,22}$&$ 78_{71,85}$&$ 11_{10,12}$&${\bf 86_{81,93}}$\\
& & 15 &$ 2_{2,8}$&$ 14_{11,71}$&$ 10_{1,10}$&$ 76_{0,85}$&$ 23_{13,30}$&$ 73_{66,82}$&$ 28_{10,30}$&$ 69_{62,89}$&$ 13_{10,15}$&$ 77_{70,81}$&$ 11_{10,16}$&${\bf 80_{72,89}}$\\
& 15 & 5 & $ 13_{11,15}$&$ 80_{70,88}$&$ 4_{1,15}$&$ 26_{0,90}$&$ 15_{15,17}$&${\bf 89_{85,95}}$&$ 15_{14,18}$&$ 89_{84,95}$&$ 15_{14,15}$&$ 85_{76,91}$&$ 16_{15,19}$&$ 88_{81,94}$\\
& & 10 & $ 11_{2,14}$&$ 66_{10,86}$&$ 1_{1,15}$&$ 0_{0,89}$&$ 19_{15,24}$&${\bf 85_{79,91}}$&$ 15_{14,16}$&${\bf 87_{77,91}}$&$ 20_{16,23}$&$ 83_{77,88}$&$ 17_{15,19}$&${\bf 86_{78,90}}$\\
& & 15 & $ 7_{2,14}$&$ 42_{8,72}$&$ 11_{1,15}$&$ 0_{0,82}$&$ 29_{26,30}$&$ 74_{69,78}$&$ 26_{15,29}$&${\bf 75_{72,82}}$&$ 19_{16,22}$&${\bf 77_{72,81}}$&$ 16_{9,21}$&${\bf 77_{51,82}}$\\
\hline
Within & 5 & 5 & $ 5_{3,5}$&${\bf 98_{60,100}}$&$ 5_{5,6}$&${\bf 98_{87,99}}$&$ 5_{5,5}$&${\bf 98_{95,100}}$&$ 24_{5,30}$&$ 34_{26,98}$&$ 15_{10,25}$&$ 48_{31,68}$&$ 5_{5,5}$&${\bf 98_{94,100}}$\\
cluster &  & 10 & $ 4_{2,5}$&$ 76_{33,97}$&$ 5_{5,8}$&${\bf 94_{74,98}}$&$ 5_{5,5}$&${\bf 95_{92,98}}$&$ 28_{24,30}$&$ 27_{25,33}$&$ 20_{14,25}$&$ 38_{29,48}$&$ 5_{5,5}$&${\bf 95_{90,98}}$\\
shape & & 15 & $ 2_{2,5}$&$ 30_{21,88}$&$ 5_{5,5}$&${\bf 86_{83,92}}$&$ 5_{5,5}$&${\bf 87_{83,92}}$&$ 29_{26,30}$&$ 24_{22,28}$&$ 18_{15,22}$&$ 34_{29,43}$&$ 5_{5,5}$&${\bf 86_{81,92}}$\\
varies& 10 & 5 & $ 9_{7,10}$&$ 85_{67,97}$&$ 10_{1,10}$&$ 94_{0,97}$&$ 10_{10,10}$&${\bf 95_{92,97}}$&$ 27_{10,30}$&$ 55_{50,94}$&$ 22_{17,25}$&$ 65_{56,79}$&$ 10_{10,11}$&$ 94_{90,96}$\\
& & 10 & $ 7_{2,10}$&$ 67_{13,94}$&$ 10_{1,10}$&$ 90_{0,94}$&$ 10_{10,10}$&${\bf 92_{88,95}}$&$ 28_{21,30}$&$ 51_{46,63}$&$ 21_{17,25}$&$ 63_{52,72}$&$ 11_{10,11}$&$ 91_{86,93}$\\
& & 15 & $ 2_{2,8}$&$ 13_{11,70}$&$ 6_{1,10}$&$ 46_{0,87}$&$ 10_{10,11}$&${\bf 84_{80,88}}$&$ 28_{25,30}$&$ 46_{42,51}$&$ 19_{15,25}$&$ 61_{52,72}$&$ 11_{10,11}$&${\bf 83_{81,87}}$\\
& 15 & 5 & $ 14_{10,15}$&$ 83_{63,90}$&$ 6_{1,15}$&$ 41_{0,95}$&$ 15_{14,15}$&${\bf 91_{86,96}}$&$ 15_{14,29}$&$ 90_{70,95}$&$ 22_{16,26}$&$ 80_{72,88}$&$ 16_{15,18}$&${\bf 90_{85,95}}$\\
& & 10 & $ 11_{2,14}$&$ 67_{9,84}$&$ 1_{1,15}$&$ 0_{0,91}$&$ 15_{15,15}$&${\bf 91_{86,94}}$&$ 15_{15,30}$&$ 87_{64,94}$&$ 23_{18,26}$&$ 76_{69,85}$&$ 16_{15,17}$&$ 88_{85,92}$\\
& & 15 & $ 2_{2,12}$&$ 8_{6,66}$&$ 1_{1,15}$&$ 0_{0,80}$&$ 15_{15,16}$&${\bf 82_{78,86}}$&$ 28_{25,29}$&$ 62_{57,65}$&$ 20_{17,23}$&$ 72_{66,79}$&$ 16_{15,17}$&$ 80_{76,85}$
\end{tabular}
}
%*** Because of the running time to compute the gap statistic, the isolet data set was projected into a 100 dimensional subspace defined by a random rotation of the first 100 principal components. All methods were applied to this reduced version of the data set. The principal components were rotated to avoid standardisation of the columns removing important information.
\end{table*}

\begin{table*}
\centering
\caption{Results from simulated non-Gaussian mixture data sets. The Median of the number of clusters selected by each method ($\hat k$) and corresponding adjusted Rand index (ARI) are reported. Subscripts show the 10$^{th}$ and 90$^{th}$ centiles. The quantiles are based on the results from 30 data sets generated for each simulation set-up. Highest performances for each scenario are highlighted , as are those which are not significantly different from the highest based on a paired Wilcoxon signed rank test with $p$-value threshold of 0.01.
\label{tb:non_gauss_sims}
}
\scalebox{0.7}{
\begin{tabular}{lcccccccccccccc}
&&& \mcl{2}{fK} & \mcl{2}{Gap} & \mcl{2}{Silh.} & \mcl{2}{Jump} &  \mcl{2}{BIC} & \mcl{2}{BIC$_{\widehat{edf}}$}   \\
Simulation & k & d & $\hat{k}$ & ARI & $\hat{k}$ & ARI  & $\hat{k}$ & ARI& $\hat{k}$ & ARI   & $\hat{k}$ & ARI  & $\hat{k}$ & ARI   \\
\hline
Long & 5 & 5 & $ 5_{2,5}$&$ 92_{36,95}$&$ 5_{1,5}$&$ 92_{0,94}$&$ 6_{5,6}$&${\bf 93_{90,95}}$&$ 5_{5,6}$&${\bf 93_{90,95}}$&$ 21_{14,26}$&$ 41_{34,58}$&$ 5_{5,6}$&${\bf 93_{90,95}}$\\
tails ($t_3$) & & 10 & $ 3_{2,5}$&$ 52_{29,86}$&$ 1_{1,5}$&$ 0_{0,90}$&$ 6_{5,8}$&$ 86_{83,90}$&$ 27_{5,30}$&$ 36_{32,87}$&$ 9_{7,14}$&$ 77_{62,85}$&$ 5_{5,7}$&${\bf 87_{84,90}}$\\
& & 15 & $ 2_{2,4}$&$ 26_{19,64}$&$ 1_{1,1}$&$ 0_{0,8}$&$ 6_{2,7}$&${\bf 78_{23,81}}$&$ 28_{24,30}$&$ 34_{30,39}$&$ 13_{9,18}$&$ 61_{43,69}$&$ 5_{1,22}$&$59_{0,81}$\\
& 10 & 5 & $ 8_{2,10}$&$ 72_{17,90}$&$ 1_{1,10}$&$ 0_{0,90}$&$ 11_{10,12}$&${\bf 90_{86,92}}$&$ 10_{10,12}$&${\bf 89_{78,92}}$&$ 18_{12,23}$&$ 77_{70,89}$&$ 11_{10,12}$&${\bf 90_{87,92}}$\\
& & 10 & $ 3_{2,9}$&$ 15_{13,74}$&$ 1_{1,2}$&$ 0_{0,8}$&$ 13_{11,14}$&${\bf 82_{80,86}}$&$ 24_{10,30}$&$ 68_{60,85}$&$ 12_{11,14}$&${\bf 83_{79,86}}$&$ 11_{11,13}$&${\bf 82_{80,86}}$\\
& & 15 & $ 2_{2,10}$&$ 12_{10,70}$&$ 1_{1,1}$&$ 0_{0,0}$&$ 15_{11,17}$&${\bf 72_{67,75}}$&$ 28_{24,30}$&$ 59_{53,66}$&$ 15_{12,17}$&${\bf 71_{67,75}}$&$ 11_{10,13}$&${\bf 73_{68,75}}$\\
& 15 & 5 & $ 11_{2,14}$&$ 64_{11,82}$&$ 1_{1,14}$&$ 0_{0,83}$&$ 16_{14,18}$&${\bf 85_{81,89}}$&$ 15_{14,17}$&${\bf 85_{78,89}}$&$ 19_{16,26}$&$ 81_{77,88}$&$ 16_{13,18}$&${\bf 85_{77,88}}$\\
& & 10 & $ 11_{2,15}$&$ 60_{9,78}$&$ 1_{1,1}$&$ 0_{0,0}$&$ 19_{16,22}$&${\bf 80_{76,83}}$&$ 23_{16,28}$&${\bf 78_{72,83}}$&$ 18_{16,22}$&${\bf 80_{75,84}}$&$ 17_{10,18}$&${\bf 80_{48,83}}$\\
& & 15 & $ 13_{2,17}$&$ 56_{7,65}$&$ 1_{1,1}$&$ 0_{0,0}$&$ 23_{19,25}$&${\bf 68_{66,74}}$&$ 28_{23,30}$&$ 66_{62,70}$&$ 19_{16,22}$&${\bf 69_{65,74}}$&$ 18_{15,20}$&${\bf 69_{63,72}}$\\
\hline
Uniform & 5 & 5 & $ 5_{2,5}$&$ 100_{38,100}$&$ 5_{5,5}$&${\bf 100_{98,100}}$&$ 5_{5,5}$&${\bf 100_{98,100}}$&$ 5_{5,5}$&${\bf 100_{99,100}}$&$ 25_{17,30}$&$ 30_{25,39}$&$ 5_{5,5}$&${\bf 100_{99,100}}$\\
clusters & & 10 & $ 4_{2,5}$&$ 76_{32,97}$&$ 5_{1,5}$&${\bf 96_{0,98}}$&$ 5_{5,5}$&${\bf 96_{94,98}}$&$ 5_{5,5}$&${\bf 96_{95,98}}$&$ 18_{15,22}$&$ 38_{31,45}$&$ 5_{5,5}$&${\bf 96_{95,98}}$\\
& & 15 & $ 2_{2,5}$&$ 28_{20,84}$&$ 5_{1,5}$&${\bf 86_{0,88}}$&$ 5_{5,5}$&${\bf 86_{84,88}}$&$ 5_{5,29}$&${\bf 86_{20,88}}$&$ 13_{11,14}$&$ 43_{39,48}$&$ 5_{1,5}$&$ 85_{0,88}$\\
& 10 & 5 & $ 8_{2,10}$&$ 80_{17,99}$&$ 10_{1,10}$&$ 95_{0,99}$&$ 10_{10,10}$&${\bf 99_{96,100}}$&$ 10_{10,10}$&${\bf 99_{97,100}}$&$ 21_{17,27}$&$ 65_{53,75}$&$ 10_{10,11}$&$ 97_{92,99}$\\
& & 10 & $ 2_{2,10}$&$ 16_{13,95}$&$ 1_{1,10}$&$ 0_{0,95}$&$ 10_{10,10}$&${\bf 94_{91,96}}$&$ 10_{10,10}$&${\bf 94_{91,96}}$&$ 17_{14,19}$&$ 71_{65,79}$&$ 10_{10,11}$&${\bf 93_{91,96}}$\\
& & 15 & $ 2_{2,2}$&$ 12_{10,13}$&$ 1_{1,2}$&$ 0_{0,8}$&$ 10_{10,10}$&${\bf 84_{81,86}}$&$ 10_{10,10}$&${\bf 83_{80,85}}$&$ 11_{11,13}$&$ 79_{72,84}$&$ 10_{10,10}$&${\bf 83_{79,85}}$\\
& 15 & 5 & $ 12_{2,14}$&$ 80_{13,92}$&$ 14_{1,15}$&$ 89_{0,98}$&$ 15_{14,15}$&$ 97_{89,98}$&$ 15_{14,15}$&${\bf 96_{90,98}}$&$ 23_{20,29}$&$ 82_{68,86}$&$ 15_{15,16}$&${\bf 95_{91,98}}$\\
& & 10 & $ 8_{2,14}$&$ 48_{9,89}$&$ 1_{1,15}$&$ 0_{0,93}$&$ 15_{15,15}$&${\bf 93_{90,95}}$&$ 15_{14,15}$&${\bf 93_{89,94}}$&$ 17_{16,17}$&$ 90_{86,92}$&$ 16_{15,16}$&${\bf 92_{88,94}}$\\
& & 15 & $ 2_{2,13}$&$ 8_{7,71}$&$ 1_{1,1}$&$ 0_{0,0}$&$ 15_{15,15}$&${\bf 81_{78,84}}$&$ 15_{14,15}$&${\bf 81_{75,84}}$&$ 15_{15,15}$&${\bf 81_{78,85}}$&$ 15_{15,16}$&$ 80_{76,84}$\\
\hline
Non- & 5 & 5 & $ 5_{4,5}$&${\bf 82_{68,91}}$&$ 9_{1,11}$&$ 68_{0,81}$&$ 6_{5,8}$&${\bf 81_{71,89}}$&$ 26_{22,30}$&$ 31_{28,44}$&$ 22_{16,25}$&$ 37_{31,56}$&$ 8_{6,11}$&$ 76_{66,85}$\\
convex & & 10 & $ 4_{2,5}$&$ 50_{24,86}$&$ 14_{11,18}$&$ 49_{34,59}$&$ 9_{6,12}$&${\bf 68_{53,86}}$&$ 28_{26,30}$&$ 27_{24,30}$&$ 22_{17,30}$&$ 31_{24,45}$&$ 9_{5,12}$&${\bf 62_{42,81}}$\\
clusters & & 15 & $ 2_{2,4}$&$ 31_{23,60}$&$ 15_{1,18}$&$ 42_{0,55}$&$ 12_{6,16}$&${\bf 50_{38,85}}$&$ 29_{28,30}$&$ 25_{23,28}$&$ 23_{16,29}$&$ 30_{24,43}$&$ 9_{4,16}$&${\bf 57_{43,81}}$\\
& 10 & 5 & $ 10_{9,10}$&${\bf 90_{85,95}}$&$ 10_{1,11}$&$ 89_{0,95}$&$ 10_{10,11}$&${\bf 92_{85,95}}$&$ 10_{10,28}$&$ 85_{57,95}$&$ 19_{15,23}$&$ 74_{63,83}$&$ 12_{11,14}$&$ 88_{82,93}$\\
& & 10 & $ 9_{8,10}$&$ 83_{69,95}$&$ 13_{1,16}$&$ 79_{0,87}$&$ 10_{10,14}$&${\bf 91_{81,96}}$&$ 28_{26,30}$&$ 57_{51,60}$&$ 22_{16,30}$&$ 65_{51,81}$&$ 14_{12,21}$&$ 85_{66,91}$\\
& & 15 & $ 9_{8,10}$&$ 82_{65,91}$&$ 16_{10,19}$&$ 74_{49,85}$&$ 12_{10,13}$&${\bf 88_{78,96}}$&$ 29_{26,30}$&$ 51_{48,56}$&$ 22_{17,27}$&$ 61_{52,72}$&$ 13_{11,17}$&$ 81_{64,91}$\\
& 15 & 5 & $ 14_{12,15}$&$ 88_{79,94}$&$ 15_{1,16}$&$ 90_{0,96}$&$ 15_{15,16}$&${\bf 93_{90,97}}$&$ 16_{15,27}$&$ 92_{71,97}$&$ 20_{16,25}$&$ 85_{76,92}$&$ 17_{16,19}$&$ 91_{84,94}$\\
& & 10 & $ 14_{13,15}$&$ 93_{84,98}$&$ 15_{12,18}$&$ 94_{74,99}$&$ 15_{15,17}$&${\bf 97_{91,99}}$&$ 28_{26,30}$&$ 76_{70,80}$&$ 20_{15,26}$&$ 89_{75,95}$&$ 18_{15,21}$&$ 92_{86,97}$\\
& & 15 & $ 15_{12,15}$&$ 92_{79,99}$&$ 16_{1,17}$&$ 94_{0,97}$&$ 15_{15,16}$&${\bf 98_{95,99}}$&$ 28_{27,30}$&$ 73_{70,79}$&$ 23_{17,26}$&$ 83_{74,95}$&$ 17_{16,19}$&$ 94_{88,97}$
\end{tabular}
}
%*** Because of the running time to compute the gap statistic, the isolet data set was projected into a 100 dimensional subspace defined by a random rotation of the first 100 principal components. All methods were applied to this reduced version of the data set. The principal components were rotated to avoid standardisation of the columns removing important information.
\end{table*}

\subsection{Public Benchmark Data}

This section presents briefly on results from experiments using a large collection of 28 publicly available data sets associated with real applications\footnote{The Synth data set is, as far as the author is aware, the only simulated data set in this collection. This data set is a popular time-series clustering data set based on short length control-chart simulations.} from diverse fields. These are popular benchmark data sets taken from the UCI machine learning repository~\citep{UCI}, with the exception of the Yeast\footnote{\url{https://genome-www.stanford.edu/cellcycle/}} and Phoneme\footnote{\url{https://web.stanford.edu/~hastie/ElemStatLearn/}} data sets. These data sets were chosen since ground-truth label sets are available, which can be used for validation and comparison of clustering solutions. All data sets were standardised to have unit variance in every dimension before applying any clustering.

\begin{figure}
\centering
\caption{Boxplots of normalised regret (when compared with the ideal performance). Mean normalised regret is indicated in each case by a red dot.\label{fig:boxplots}}
\includegraphics[width = 12cm]{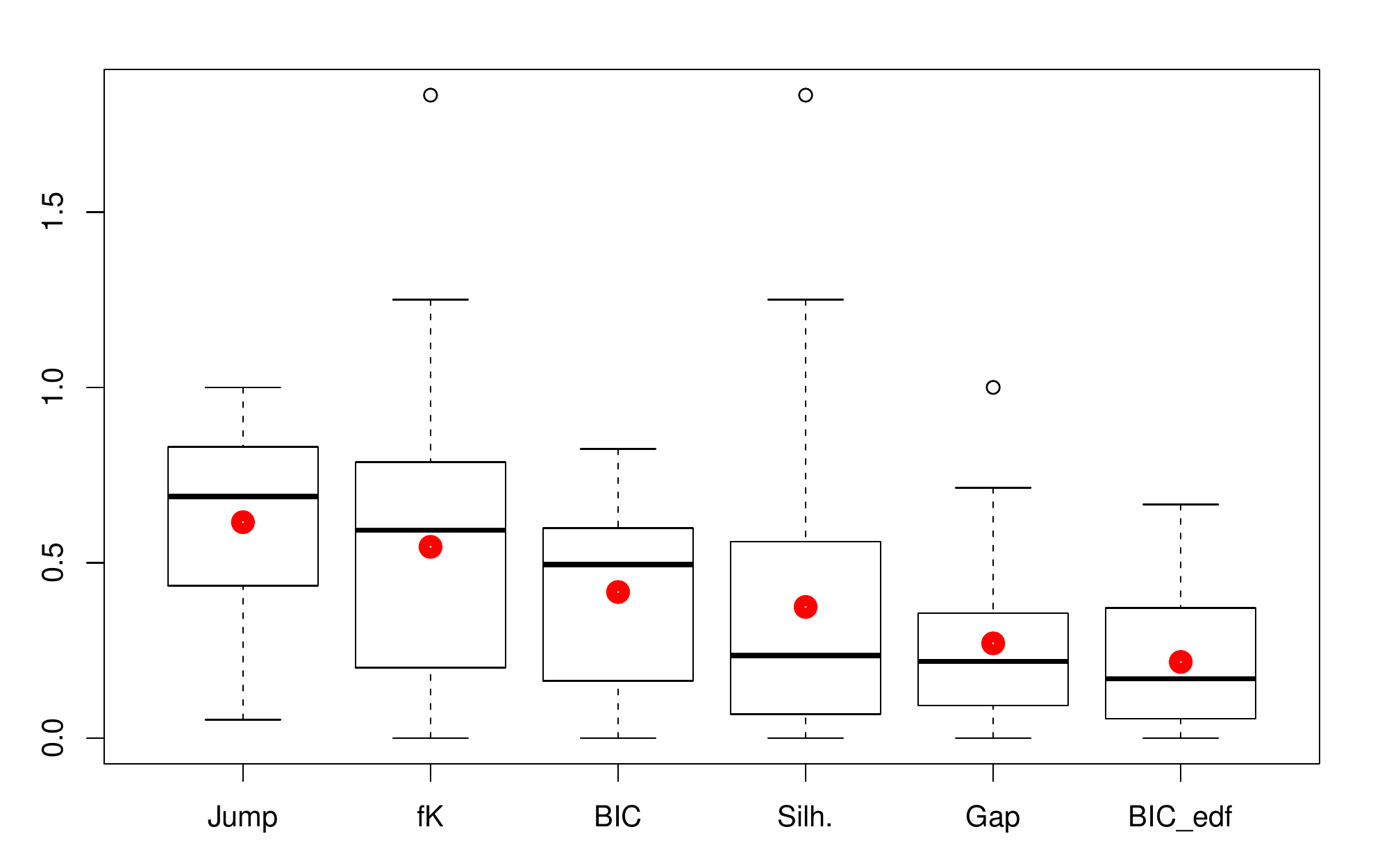}
\end{figure}

Table~\ref{tb:real} shows the results of these experiments. The numbers in brackets indicate the true number of clusters, $k$. For each method the selected number of clusters, $\hat k$, and the adjusted Rand index are reported\footnote{Two of the data sets offer multiple ``ground truth'' label sets. The table shows the average performance of each method over the different label sets.}. For each data set we have also included the ``Ideal'' $k$-means solution, which corresponds with the solution that attains the highest ARI value. We find this to be pertinent since when the data distribution deviates substantially from the $k$-means assumptions it may be that the best $k$-means solution does not contain the same number of clusters as the ground truth. Furthermore, although it is unlikely that there exists a method which will reliably select the ideal solution, it is also very likely that there exists, theoretically, a method which performs better than any of the methods considered herein. Comparing with the ideal performance therefore gives a bound on how much better it is possible to perform with any model selection technique for $k$-means. The ideal performance also gives us some indication of the difficulty of the clustering problem. Two immediate take-aways from the table are that the fK method
selected two clusters in almost all cases, while the Jump Statistic dramatically over-estimated the number of clusters in all but a few instances. The BIC with na{\"i}ve setting of the degrees of freedom also over-esimates the number of clusters considerably in general, but not by so large a margin as the Jump Statistic. The Silhouette Index, Gap Statistic and the BIC with the effective degrees of freedom all perform quite consistently well. To better illustrate the overall performance of the methods on these data sets, the results of Table~\ref{tb:real} are summarised in Figure~\ref{fig:boxplots}. The figure shows boxplots of the ARI performance regret, when compared to the ideal performance, normalised for difficulty. That is, for a method $M$ and data set $\mathbf{X}$, the normalised regret is given by
\begin{align*}
\frac{\mbox{ARI(Ideal(}\mathbf{X})) - \mbox{ARI(}M(\X))}{\mbox{ARI(Ideal(}\mathbf{X}))}.
\end{align*}
The figure also shows the mean of the normalised regret for each method, indicated by a red dot. Here we see that the Gap Statistic and the BIC using the proposed estimate of effective degrees of freedom perform substantially better than the other methods, in general. While the Silhouette Index yields a similar median performance, its instances of poor performance are considerably worse than those of the Gap and proposed BIC variant.
%
%Finally, summaries of the contents of the table are also included. These summaries include the number of times each method achieved the highest performance, as well as the averages of absolute and relative regret when compared with the Ideal solution.
%

Given the variety and number of the data sets used in these experiments, there is strong evidence that the proposed estimation procedure for the effective degrees of freedom leads to selection of models which enjoy very strong performance when compared with existing techniques.

\begin{table*}
\centering
\caption{Results from publicly available benchmark data sets. Number of clusters selected by each method ($\hat k$) and corresponding adjusted Rand index (ARI) are reported.\label{tb:real}}
\scalebox{0.9}{
\begin{tabular}{lcccccccccccccc}
& \mcl{2}{fK} & \mcl{2}{Gap} & \mcl{2}{Silh.} & \mcl{2}{Jump} &  \mcl{2}{BIC} & \mcl{2}{BIC$_{\widehat{edf}}$} & \mcl{2}{Ideal}  \\
Data set $(k)$& $\hat{k}$ & ARI & $\hat{k}$ & ARI  & $\hat{k}$ & ARI& $\hat{k}$ & ARI & $\hat{k}$ & ARI   & $\hat{k}$ & ARI  & $\hat{k}$ & ARI  \\
\hline
 Wine  (3)    & 2 & 0.37 & 3 & {\bf 0.9} & 3 & {\bf 0.9} & 30 & 0.13 & 11 & 0.35 & 3 & {\bf 0.9} &   3    & 0.9	\\
 Seeds  (3)    & 2 & 0.48 & 3 & {\bf 0.77} & 2 & 0.48 & 29 & 0.12 & 17 & 0.2 & 3 & {\bf 0.77} &   3    & 0.77	\\
 Ionosphere  (2)    & 2 & 0.17 & 8 & 0.17 & 4 & {\bf 0.28} & 30 & 0.11 & 12 & 0.12 & 4 & {\bf 0.28} &   3    & 0.29	\\
 Votes  (2)    & 2 & {\bf 0.57} & 7 & 0.21 & 2 & {\bf 0.57} & 29 & 0.06 & 14 & 0.1 & 4 & 0.32 &   2    & 0.57	\\
 Iris  (3)    & 2 & 0.57 & 3 & {\bf 0.62} & 2 & 0.57 & 27 & 0.14 & 14 & 0.3 & 3 & {\bf 0.62} &   3    & 0.62	\\
 Libras  (15)    & 2 & 0.07 & 13 & 0.31 & 18 & 0.31 & 29 & 0.29 & 16 & {\bf 0.32} & 16 & {\bf 0.32} &   20    & 0.34	\\
 Heart  (2)    & 2 & {\bf 0.34} & 2 & {\bf 0.34} & 5 & 0.29 & 29 & 0.04 & 5 & 0.29 & 5 & 0.29 &   2    & 0.34	\\
 Glass  (6)    & 2 & 0.19 & 9 & 0.17 & 2 & 0.19 & 29 & 0.13 & 13 & {\bf 0.24} & 4 & 0.2 &   5    & 0.24	\\
 Mammography  (2)    & 2 & {\bf 0.39} & 3 & 0.31 & 3 & 0.31 & 25 & 0.05 & 11 & 0.13 & 4 & 0.31 &   2    & 0.39	\\
 Parkinsons  (2)    & 2 & -0.1 & 7 & {\bf 0.07} & 2 & -0.1 & 30 & 0.03 & 12 & 0.05 & 10 & 0.04 &   6    & 0.12	\\
 Yeast  (5)    & 2 & {\bf 0.42} & 8 & 0.4 & 2 & {\bf 0.42} & 29 & 0.14 & 10 & 0.39 & 12 & 0.36 &   4    & 0.57	\\
 Forest  (4)    & 2 & 0.18 & 5 & {\bf 0.39} & 2 & 0.18 & 30 & 0.15 & 19 & 0.2 & 12 & 0.28 &   4    & 0.45	\\
 Breast Cancer  (2)    & 2 & {\bf 0.82} & 9 & 0.38 & 2 & {\bf 0.82} & 30 & 0.15 & 17 & 0.34 & 4 & 0.76 &   2    & 0.82	\\
 Dermatology  (6)    & 2 & 0.21 & 6 & {\bf 0.7} & 3 & 0.57 & 28 & 0.26 & 9 & 0.65 & 6 & {\bf 0.7} &   5    & 0.84	\\
 Synth  (6)    & 2 & 0.27 & 8 & {\bf 0.67} & 2 & 0.27 & 30 & 0.35 & 10 & 0.65 & 10 & 0.65 &   8    & 0.67	\\
 Soy Bean  (19)    & 2 & 0.05 & 16 & {\bf 0.43} & 2 & 0.05 & 30 & 0.42 & 16 & {\bf 0.43} & 16 & {\bf 0.43} &   18    & 0.56	\\
 Olive Oil  (3/9)    & 2 & 0.4 & 10 & 0.49 & 5 & {\bf 0.67} & 30 & 0.19 & 18 & 0.17 & 9 & 0.5 &   5    & 0.67	\\
 Bank  (2)    & 2 & 0.01 & 3 & 0.06 & 18 & {\bf 0.1} & 26 & 0.09 & 24 & 0.09 & 21 & {\bf 0.1} &   5    & 0.21	\\
 Optidigits  (10)    & 2 & 0.13 & 17 & 0.57 & 20 & 0.6 & 30 & 0.47 & 18 & {\bf 0.65} & 18 & {\bf 0.65} &   18    & 0.65	\\
 Image Seg  (7)    & 2 & 0.17 & 14 & 0.46 & 6 & {\bf 0.48} & 28 & 0.3 & 14 & 0.46 & 14 & 0.46 &   9    & 0.51	\\
 MF Digits  (10)    & 2 & 0.15 & 18 & 0.62 & 9 & {\bf 0.65} & 1 & 0 & 20 & 0.59 & 20 & 0.59 &   11    & 0.68	\\
 Satellite  (6)    & 3 & 0.29 & 12 & {\bf 0.41} & 3 & 0.29 & 30 & 0.25 & 16 & 0.35 & 16 & 0.35 &   7    & 0.56	\\
 Texture  (11)    & 2 & 0.11 & 23 & {\bf 0.41} & 2 & 0.11 & 30 & {\bf 0.41} & 30 & {\bf 0.41} & 30 & {\bf 0.41} &   11    & 0.5	\\
 Pen Digits  (10)    & 2 & 0.13 & 30 & 0.45 & 8 & 0.45 & 28 & 0.46 & 30 & 0.45 & 21 & {\bf 0.53} &   14    & 0.64	\\
 Phoneme  (5)    & 2 & 0.16 & 11 & {\bf 0.45} & 2 & 0.16 & 1 & 0 & 21 & 0.28 & 21 & 0.28 &   5    & 0.64	\\
 Frogs  (4/8/10)    & 2 & 0.46 & 17 & 0.21 & 3 & {\bf 0.5} & 25 & 0.14 & 17 & 0.15 & 15 & 0.24 &   4    & 0.57	\\
 Auto  (3)    & 2 & -0.04 & 4 & {\bf 0.13} & 2 & -0.04 & 26 & 0.05 & 24 & 0.03 & 4 & {\bf 0.13} &   5    & 0.16	\\
 Yeast UCI  (10)    & 7 & {\bf 0.19} & 1 & 0 & 6 & 0.11 & 9 & 0.18 & 4 & 0.1 & 9 & 0.18 &   7    & 0.19	\\
 \end{tabular}
}
%*** Because of the running time to compute the gap statistic, the isolet data set was projected into a 100 dimensional subspace defined by a random rotation of the first 100 principal components. All methods were applied to this reduced version of the data set. The principal components were rotated to avoid standardisation of the columns removing important information.
\end{table*}

%%%%%%%%%%%%%%%%%%%%%%%%%%%%%%%%%%%%

\iffalse

\subsection{A Brief Comment on Running Times}

One of the additional benefits offered by the proposed approach, which has not been a focus of this piece of work, is that it is considerbly more computationally efficient than certain of the other popular and successful selection techniques. While not as computationally efficient as those based on simple and direct transformations of the $k$-means objective, such as the fK method, the jump statistic and the na{\"i}ve BIC, its computational complexity is linear in both the number of data and the number of dimensions (i.e., the same as that of the $k$-means clustering component itself) and also requires minimal additional clustering\footnote{Recall that in order to compute the degrees of freedom estimate, we use an additional $k$-means solution with $k_{max} + 1$ clusters.}. On the other hand, to compute the silhouette index the entire collection of pairwise dissimilarities between data are needed, resulting in computational complexity which is quadratic in the number of data; while methods based on re-sampling or Monte Carlo, like the Gap statistic, require a very large number of additional runs of the $k$-means algorithm. 

\fi

%%%%%%%%%%%%%%%%%%%%%%%%%%%%%%%%%%%%%%%%

%\begin{figure}
%\centering
%\includegraphics[width = 6cm]{boxplots.pdf}
%\end{figure}

\section{Discussion} \label{sec:discussion}

This work investigated the effective degrees of freedom in the $k$-means model. We argued that the degrees of freedom estimate based on the number of explicitly estimated parameters is an inappropriate pairing with the so-called classification likelihood for performing model selection for $k$-means. This is because the classification likelihood assumes the clustering assignment forms part of the estimation, but this added estimation is not accounted for in the model dimension. The proposed formulation accommodates the uncertainty of the class assignments in the degrees of freedom, where an extension of Stein's lemma showed that these uncertainties are appropriately accommodated by considering the size and location of the discontinuities in the $k$-means model, which correspond precisely to the reassignments of points to different clusters. Evaluating the new degrees of freedom expression is challenging, however a few simplifications allowed us to approximate this value in practice. The approximation was validated through model selection within the Bayesian Information Criterion.
%
% used the notion of effective degrees of freedom to obtain an alternative to the standard application of the Bayesian Information Criterion for model selection in $k$-means clustering. An extension of Stein's lemma made it possible to approximate the effective degrees of freedom. %A thorough simulation study illustrated the effectiveness and robustness of this approach in model selection for the $k$-means method. 
Experiments using simulated data, as well as a large collection of publicly available benchmark data sets suggest that this approach is competitive with popular existing methods for model selection in $k$-means clustering.

%\appendix
\section*{Proofs} 

\subsection*{Proof of Lemma 1}

Let $X\sim N(\mu, \sigma^2)$ and consider any $g:\R\to\R$ which is Lipschitz on $(-\infty, \delta)$ and $(\delta, \infty)$ for some $\delta \in \R$. For each $\epsilon > 0$ define

$$
g_\epsilon(x) = \left\{\begin{array}{ll}
g(x), & x \not \in \B_\epsilon(\delta)\\
g(\delta-\epsilon)+[g(\delta+\epsilon)-g(\delta-\epsilon)]\frac{x-(\delta-\epsilon)}{2\epsilon}, & x \in \B_\epsilon(\delta),
\end{array}\right.
$$

where $\B_\epsilon(\delta) = (\delta-\epsilon, \delta+\epsilon)$. Then $g_\epsilon$ is Lipschitz by construction and so by~\cite[Lemma 3.2]{CandesSURE} we know $g_\epsilon$ is almost differentiable and $E[g_{\epsilon}^\prime(X)^2]<\infty$, and so by \citep[Lemma 2]{SteinSURE} we have

$$
\frac{1}{\sigma^2}E[(X-\mu)g_\epsilon(X)] = E[g^\prime_\epsilon(X)].
$$

But
\begin{align*}
E[g^\prime_\epsilon(X)] =& E\left[g_\epsilon^\prime(X)|X\not \in \B_\epsilon(\delta)\right]P(X\not \in \B_\epsilon(\delta))+ E\left[g_\epsilon^\prime(X)|X \in \B_\epsilon(\delta)\right]P(X \in \B_\epsilon(\delta))\\
=&E\left[g^\prime(X)|X\not \in \B_\epsilon(\delta)\right]P(X\not \in \B_\epsilon(\delta)) + \frac{g(\delta+\epsilon)-g(\delta-\epsilon)}{2\epsilon}P(X \in \B_\epsilon(\delta)).
\end{align*}
Taking the limit as $\epsilon \to 0^+$ gives
\begin{align*}
\frac{1}{\sigma^2}E[(X-&\mu)g(X)] = E[g^\prime(X)]+ \left(\lim_{\gamma\downarrow\delta}g(\gamma)-\lim_{\gamma\uparrow\delta}g(\gamma)\right)\frac{1}{\sqrt{2\pi}\sigma}e^{\frac{-1}{2\sigma^2}(\delta - \mu)^2},
\end{align*}
as required. The extension to any $g$ with finitely many such discontinuity points arises from a very simple induction.

% Notice also that since $\bigcup_{i=1}^n\{f(\Y)_{i\_}\} \subset \mbox{\bf conv}\left(\bigcup_{j=1}^n \{\Y_{j\_}\}\right)$, and $\Y$ is normally distributed, we know that

%

%\begin{align*}

%E\left[\left(\sum_{\substack{\delta: \Y_{i,j}+\delta \in \\ \mathcal{D}(\{\Y_{k,l}\}_{(k,l)\not = (i,j)})}}\phi\left(\frac{\Y_{i,j}+\delta - \mmu_{i,j}}{\sigma}\right)\lim_{\gamma \downarrow\uparrow \delta}f(\Y+\gamma \e_{i,j})_{i,j}\right)^2\right] < \infty,

%\end{align*}

%

%and hence both terms on the right hand side of Eq.~(\ref{eq:lem1}) exist and are finite (with the first coming from~\citep[Theorem ]{}).

We therefore have for any $i, j$, that
\begin{align*}
\frac{1}{\sigma^2}E\bigg[(\X_{i,j} - \mmu_{i,j})f&(\X)_{i,j}\bigg| \{\X_{k,l}\}_{(k,l)\not = (i,j)}\bigg] = E\left[\frac{\partial}{\partial X_{i,j}}f(\X)_{i,j}\bigg| \{\X_{k,l}\}_{(k,l)\not = (i,j)}\right]\\
&+ \sum_{\substack{\delta: \X_{i,j}+\delta \in \\ \mathcal{D}(\{\X_{k,l}\}_{(k,l)\not = (i,j)})}}\frac{1}{\sigma}\phi\left(\frac{\X_{i,j}+\delta - \mmu_{i,j}}{\sigma}\right)\lim_{\gamma \downarrow\uparrow \delta}f(\X+\gamma \e_{i,j})_{i,j}
\end{align*}
The result follows from the law of total expectation.\hfill $\square$

\subsection*{Proof of Lemma 2}

Notice that the discontinuities in $\M(\X)_{i,j}$ can occur only when there is a change in the assignment of one of the observations. If this occurs at the point $\X + \delta \e_{i,j}$, then it is straightforward to show that 
$$
|\lim_{\gamma \downarrow\uparrow \delta}\M(\X+\gamma \e_{i,j})_{i,j}| \leq \diam{\X} + C|\delta|,
$$
where Diam$(\X)$ is the diameter of the rows of $\X$ and $C$ is a constant independent of $\X$. There are also clearly finitely many such discontinuities since there are finitely many cluster solutions arising from $n$ data, i.e., 
$$
| \mathcal{D}(\{\X_{k,l}\}_{(k,l)\not = (i,j)})| \leq A,
$$
for some constant $A$ independent of $i, j, \X$. Furthermore $|\M(\X + \gamma \e_{i,j})_{i,j} - \M(\X)_{i,j}| \leq \gamma$ as long as all cluster assignments remain the same, and hence $\M(\X + \gamma \e_{i,j})_{i,j}$ is Lipschitz as a function of $\gamma$ between points of discontinuity. Finally,
\begin{align*}
&E\Bigg[\Bigg|\sum_{\substack{\delta: \X_{i,j}+\delta \in \\ \mathcal{D}(\{\X_{k,l}\}_{(k,l)\not = (i,j)})}}\phi\left(\frac{\X_{i,j}+\delta - \mmu_{i,j}}{\sigma}\right)\lim_{\gamma \downarrow\uparrow \delta}\M(\X+\gamma \e_{i,j})_{i,j}\Bigg|\Bigg]\\
&\leq E\Bigg[\sum_{\substack{\delta: \X_{i,j}+\delta \in \\ \mathcal{D}(\{\X_{k,l}\}_{(k,l)\not = (i,j)})}}\phi\left(\frac{\X_{i,j}+\delta - \mmu_{i,j}}{\sigma}\right)(\diam{\X} + C|\delta|)\Bigg]\\
&\leq \frac{A}{\sqrt{2\pi}} \bigg(E[\diam{\X}] + CE\left[|\X_{i,j} - \mmu_{i,j}|+(\X_{i,j} - \mmu_{i,j})^2 + 4\sigma^2\right]\bigg),
\end{align*}
since $\phi((a-\delta)/\sigma)|\delta|$ is maximised by a $\delta$ satisfying $|\delta| \leq (|a|+|a^2-4\sigma^2|)/2$, and $\phi$ is bounded above by $1/\sqrt{2\pi}$. Now, the tail of the distribution of $\diam{\X}$ is similar to that of the distribution of the maximum of $n$ $\chi$ random variables with $d$ degrees of freedom. Therefore $E[\mbox{Diam}(\X)]$ is clearly finite. The second term above is clearly finite, since $\X_{i,j}-\mmu_{i,j}$ is normally distributed, and hence the expectation in Lemma 2 exists and is finite.\hfill $\square$

\bibliographystyle{plainnat}
%\bibliography{kmeans}

%% -- Appendix (if any) --------------------------------------------------------
%% - After the bibliography with page break.
%% - With proper section titles and _not_ just "Appendix".

\end{document}